%% file: main.tex
\documentclass{article}
\PassOptionsToPackage{table}{xcolor}
\usepackage{iclr2027_conference,times}
\input{math_commands}

\usepackage{hyperref}
\usepackage{url}
\input{preamble_additions}
\input{preprint_setup}

\title{One Readout, Many Repairs: Diffusion-Guided Hierarchical Search for Tool-Agent Repair}
\hypersetup{pdftitle={One Readout, Many Repairs: Diffusion-Guided Hierarchical Search for Tool-Agent Repair}}
\input{authors}

\begin{document}
\maketitle
\begin{abstract}
\input{sections/abstract}
\end{abstract}
\input{sections/introduction}
\input{sections/related_work}
\input{sections/problem}
\input{sections/method}
\input{sections/experiments}
\input{sections/discussion}
\bibliography{references}
\bibliographystyle{iclr2027_conference}
\clearpage
\appendix
\input{sections/appendix}

\end{document}

%% file: math_commands.tex
\usepackage{amsmath,amsfonts,bm}

\def\eqref#1{equation~\ref{#1}}

\def\1{\bm{1}}

\DeclareMathAlphabet{\mathsfit}{\encodingdefault}{\sfdefault}{m}{sl}
\SetMathAlphabet{\mathsfit}{bold}{\encodingdefault}{\sfdefault}{bx}{n}



%% file: preamble_additions.tex
\usepackage[T1]{fontenc}
\usepackage{amssymb,booktabs,graphicx,array}
\usepackage[table]{xcolor}
\usepackage{multirow}
\usepackage{wrapfig}
\definecolor{recommitTableHeader}{HTML}{F0F2F5}
\definecolor{recommitTableHighlight}{HTML}{E7F1FA}
\definecolor{recommitTableAccent}{HTML}{1F557A}

\hypersetup{hypertexnames=false,hidelinks}

%% file: preprint_setup.tex
\iclrfinalcopy
\usepackage{etoolbox}
\makeatletter
\patchcmd{\@maketitle}
  {\lhead{Published as a conference paper at ICLR 2027}}
  {\lhead{Preprint}}
  {}
  {\PackageError{recommit-preprint}{Unable to replace the conference header}{Check the ICLR style before generating the preprint.}}
\makeatother

%% file: authors.tex
\author{%
\begin{tabular}{@{}c@{}}
\textbf{Xiang Xia\thanks{Equal contribution.} \quad Cheng Yan\footnotemark[1] \quad Wuyang Zhang\thanks{Corresponding author.} \quad Fan Xu} \\[2pt]
\textbf{Zhijun Fan \quad Shuyuan Zhang \quad Yanyong Zhang} \\[3pt]
\normalfont University of Science and Technology of China, Hefei, China \\[2pt]
\normalfont\fontsize{8}{10}\selectfont \href{mailto:xxia@mail.ustc.edu.cn}{\nolinkurl{xxia@mail.ustc.edu.cn}} \quad
\href{mailto:yc_sa22218099@mail.ustc.edu.cn}{\nolinkurl{yc_sa22218099@mail.ustc.edu.cn}} \quad
\href{mailto:wuyangz@ustc.edu.cn}{\nolinkurl{wuyangz@ustc.edu.cn}}
\end{tabular}%
}
\hypersetup{pdfauthor={Xiang Xia, Cheng Yan, Wuyang Zhang, Fan Xu, Zhijun Fan, Shuyuan Zhang, Yanyong Zhang}}

%% file: sections/abstract.tex
Tool agents use large language models to act through external tools, yet successfully executed calls can still leave user requests unfulfilled. Tool-agent repair seeks alternative call sequences that execute successfully and fulfill the original requests. However, repair requires exploring both operation choices and their concrete realizations, making complete-sequence regeneration costly. Moreover, regeneration repeats operation selection even when failure arises from how those operations are realized. The resulting challenge is to reduce this repetition while preserving exploration of alternative operations and realizations. Therefore, we formulate repair as hierarchical search over operation supports, which we introduce as sets of permitted operation types that define reusable search regions for concrete tool-call sequences. We propose ReCommit, a training-free, diffusion-guided framework for improving tool-agent failure recovery while reducing repair computation. ReCommit amortizes operation-level proposal computation across repair trials by reusing operation-type scores from a single parallel readout of a masked diffusion language model. These scores guide search across supports, while realization search explores alternative entity bindings, arguments, and action composition within each support. Experiments on real failures across four enterprise services in the Agent-Diff benchmark show 75.9\% and 63.2\% relative recovery gains with 61.3\% and 51.3\% reductions in mean full-budget repair time at repair budgets $B=3$ and $B=13$, respectively, over the strongest evaluated 8B comparison method. ReCommit achieves a favorable recovery--cost trade-off, including in comparisons with the evaluated 32B models.

%% file: sections/introduction.tex
\section{Introduction}
\label{sec:introduction}
Tool agents enable large language models (LLMs) to interact with external applications by planning and executing tool-call sequences~\citep{toolformer,gorilla,toolllm}. These agents retrieve information and change application state to carry out multi-step tasks~\citep{react,codeact,apibank}. Such agents have been built and evaluated for web interaction~\citep{webarena}, software engineering~\citep{sweagent}, and multi-application digital workflows~\citep{appworld}. However, tool calls can execute without errors while leaving the user's request unfulfilled. Tool-agent repair seeks an alternative tool-call sequence that executes successfully and fulfills the original request. This requires exploring both operation choices and their concrete realizations through entity bindings, arguments, and action composition.

\begin{figure}[t]
    \centering
    \includegraphics[width=0.98\linewidth]{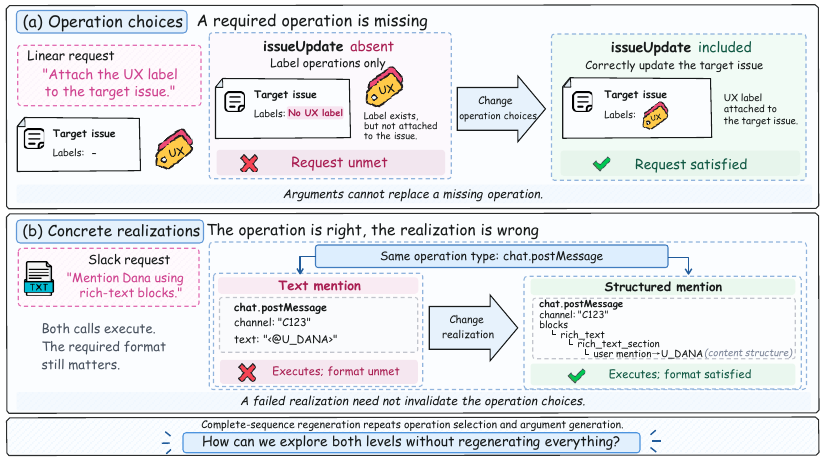}
    \vspace{-5pt}
    \caption{Two distinct repair needs from real failures in Agent-Diff~\citep{agentdiff}. (a) In Linear, the failed attempt omits \texttt{issueUpdate}, which is needed to update the target issue. (b) In Slack, both realizations use \texttt{chat.postMessage}, but only the structured user mention satisfies the requested representation. Calls are abbreviated.}
    \label{fig:motivation}
    \vspace{-10pt}
\end{figure}

Two repair cases on enterprise API tasks from the Agent-Diff benchmark~\citep{agentdiff} illustrate these needs. In the Linear issue-management task of Figure~\ref{fig:motivation}(a), calls that manipulate labels fail to attach the requested label to the target issue because the required update operation is missing. In the Slack messaging task of Figure~\ref{fig:motivation}(b), the messaging operation is already appropriate, but the user mention must be expressed in the requested structured format. The Linear case requires a different operation choice, whereas the Slack case requires a different realization of a suitable choice. These cases reveal that repair must accommodate changes at both levels, while a failed realization does not necessarily invalidate its operation choices.

\noindent\textbf{Problem.}
Feedback and reflection can help agents revise unsuccessful attempts~\citep{reflexion,critic}, but such guidance alone does not specify how to reuse operation-level proposal computation across repair trials. Repair search spans different combinations of operation types, each admitting concrete realizations with different entity bindings, arguments, and action composition. Within a limited trial budget, repair must explore alternative operation choices while retaining opportunities to test different realizations of each choice. Regenerating a complete tool-call sequence for every trial repeats operation selection even when only the realization needs to change. Fixing operation choices avoids regenerating those choices but can exclude repairs that require different operations. Under a limited computational budget, how can repair reuse operation-level proposal computation while preserving exploration of alternative operation choices and concrete realizations?

\noindent\textbf{Contribution.}
We introduce operation supports as an abstraction for tool-agent repair and formulate repair as \emph{hierarchical search over these supports and their concrete realizations}. An operation support specifies a set of permitted state-changing operation types and defines a realization region containing concrete tool-call sequences. It connects operation choices to their realizations without fixing entity bindings, arguments, or action composition. Search across supports changes the permitted operation types, whereas search within a support explores different concrete realizations. This formulation makes operation-level proposals reusable across repair trials while preserving exploration within the corresponding realization regions.

We propose ReCommit, a training-free, diffusion-guided framework for improving tool-agent failure recovery while reducing repair computation. ReCommit realizes this hierarchy through amortized support search, which reuses a shared scoring basis across repair trials, and support-conditioned realization search. A masked diffusion language model (dLLM)~\citep{llada,dream} provides parallel operation-type predictions under a shared repair context. The predictions from a single readout are pooled into inclusion scores, which are reused to rank multiple supports without additional dLLM calls. Within each support, canonical realization grounds a schema-based call structure, while expressive realization explores alternative entity bindings, arguments, and action composition. For each ranked support, the schedule attempts canonical realization followed by expressive realization before moving to the next support. Thus, reusing operation-level scores does not fix the concrete tool-call sequence. Public-error pruning further processes rejected calls using public execution observations, with any replay starting from the initial environment state.

Experiments on real tool-agent failures from the Agent-Diff benchmark~\citep{agentdiff} cover four enterprise services. At repair budgets $B=3$ and $B=13$, ReCommit improves recovery rate by 75.9\% and 63.2\%, respectively, relative to the strongest evaluated 8B comparison method, while reducing mean full-budget repair time by 61.3\% and 51.3\%. ReCommit achieves a favorable recovery--cost trade-off, including in comparisons with the evaluated 32B models. For support search, controlled comparisons show that the evaluated dLLM proposer achieves higher recovery than the autoregressive proposer with downstream realization and execution procedures fixed. For realization search, the canonical/expressive schedule achieves higher recovery than the evaluated single-mode schedules under the same trial budget. Together, these results suggest that reusing diffusion-guided operation-level proposal computation across trials while exploring alternative supports and their realizations can improve the recovery--cost trade-off in tool-agent repair.

%% file: sections/related_work.tex
\section{Related Work}
\label{sec:related}
\noindent\textbf{Tool Agents.} Research on tool agents studies how LLMs select tools, bind arguments, and coordinate calls in external environments. Toolformer~\citep{toolformer} learns API use from self-supervised examples, while Gorilla~\citep{gorilla} and ToolLLM~\citep{toolllm} improve API invocation through model adaptation and tool-oriented data. ToolkenGPT~\citep{toolken} represents tool selection with learned embeddings. ReAct~\citep{react} interleaves reasoning and actions, while CodeAct~\citep{codeact} uses executable code to compose tool interactions. Evaluation ranges from API planning and calling in API-Bank~\citep{apibank} to multi-application tasks in AppWorld~\citep{appworld} and user--agent interaction in $\tau$-bench~\citep{taubench} and $\tau^2$-Bench~\citep{tau2}. Agent-Diff~\citep{agentdiff} evaluates whether executions produce requested state changes in enterprise applications. This setting supports studying task recovery even when individual calls are accepted.

\noindent\textbf{Tool-Agent Failures and Repair.} Tool-agent failures include incorrect tool choices, argument errors, and incomplete tasks even when tool calls execute without errors. ToolMaze~\citep{toolmaze} and ToolBench-X~\citep{toolbenchx} examine recovery under injected tool faults and misleading outputs. Revision methods differ in the feedback they use: Reflexion~\citep{reflexion} produces verbal reflections from interaction feedback, Self-Refine~\citep{selfrefine} iterates self-generated feedback and revisions, and CRITIC~\citep{critic} uses tool observations to verify and correct outputs. AgentDebug~\citep{agentdebug} localizes critical errors in failed trajectories and supplies targeted feedback for new rollouts. Language Agent Tree Search~\citep{lats} combines feedback with search over action trajectories. These works use feedback to guide revision and trajectory search. ReCommit structures repair as diffusion-guided hierarchical search across operation supports and within each support, reusing operation-level proposals.

\noindent\textbf{Diffusion Language Models.} Discrete diffusion~\citep{d3pm,sedd} and masked diffusion modeling~\citep{mdlm} formulate text generation through denoising. LLaDA~\citep{llada} and Dream~\citep{dream} develop large-scale dLLMs with instruction-tuned variants, while DiffuCoder~\citep{diffucoder} studies diffusion modeling for code generation. Masked diffusion predicts tokens at multiple positions in parallel under a shared context. DiffuAgent~\citep{diffuagent} examines dLLMs both as agent backbones and as specialized modules, including tool selection. DiG-Plan~\citep{digplan} addresses early commitment in tool-graph planning by separating diffusion-based tool-set exploration from dependency prediction. ReCommit targets the recovery--cost trade-off in tool-agent repair, reusing diffusion-guided operation-level proposal computation across trials while preserving exploration of alternative supports and their realizations.

%% file: sections/problem.tex
\section{Hierarchical Formulation of Tool-Agent Repair}
\label{sec:problem}
We formalize tool-agent repair and introduce our hierarchical search formulation over operation supports and concrete realizations. The central abstraction is an operation support, which defines a reusable search region without committing to a concrete tool-call sequence.

\subsection{Tool-Agent Repair}
Given a request $u$ and initial environment state $s_0$, a tool agent produces a tool-call sequence $\pi=((e_1,\alpha_1),\ldots,(e_L,\alpha_L))$. Call $r$ invokes operation type $e_r$ with arguments $\alpha_r$, including entity bindings. Execution yields a final state $s_\pi$ and public observations $\omega_\pi$, with $(s_\pi,\omega_\pi)=\operatorname{Exec}(s_0,\pi)$. Let $g_u(s_0,s_\pi)\in\{0,1\}$ denote the benchmark task predicate evaluated on the initial and final states. In our repair setting, success requires both valid execution and fulfillment of the request:
\begin{equation}
    y(\pi;u,s_0)=\mathbf{1}\!\left[\operatorname{Valid}(\pi,\omega_\pi)\land g_u(s_0,s_\pi)=1\right],
    \label{eq:success}
\end{equation}
where $\operatorname{Valid}$ requires a nonempty, well-formed call sequence whose final execution contains no rejected calls. Execution validity alone therefore does not imply task completion.

A failure episode provides an attempt $\pi^{\mathrm{fail}}$ with $g_u(s_0,s_{\pi^{\mathrm{fail}}})=0$. Repair searches for an alternative sequence with $y=1$ using the public repair context $x$, comprising the request, failed attempt, execution observations, tool schemas, and visible state. The environment maintains the full state $s_0$, while the framework has access only to its public information. Each alternative sequence is executed from the same $s_0$. Neither the task predicate $g_u$ nor its outcomes are available to repair search.

\subsection{From Call Sequences to Operation Supports}
A complete tool-call sequence fixes both the operation types and how they are realized through entity bindings, arguments, and action composition. We introduce \emph{operation supports} to separate these decisions. Let $\mathcal{T}$ denote the service's set of state-changing operation types. We define an operation support as a subset $S\subseteq\mathcal{T}$ of permitted types, inducing the realization region $\mathcal{R}(S)=\{\pi:\operatorname{OpTypes}(\pi)\subseteq S\}$, where $\operatorname{OpTypes}(\pi)$ returns the set of state-changing operation types in $\pi$. A \emph{realization} $\pi\in\mathcal{R}(S)$ is a concrete tool-call sequence using only permitted state-changing types. Read-only calls remain available for public lookups. The support constrains operation types while leaving entity bindings, arguments, and action composition, including repeated calls and their order, open to search. A realization need not use every type in $S$, so support regions can overlap.

A support focuses realization search on selected operation types while preserving alternative entity bindings, arguments, and action composition. Failure of one realization does not rule out successful alternatives within the same support. Search can therefore retain the operation-level proposal while varying its concrete realization. A \emph{repair trial} produces one final realization after any generation and execution processing. For distinct supports $S_1,S_2,\ldots$, let $k_t$ index the support used at trial $t$, whose final realization satisfies $\pi_t\in\mathcal{R}(S_{k_t})$. Search across supports changes the permitted operation types, whereas search within a support varies their concrete realization. Across trials, operation choices remain reusable while entity bindings, arguments, and action composition vary.

\subsection{Recovery--Cost Objective}
The goal is to recover failed tasks with less computation devoted to proposing, realizing, and executing repairs. Let the repair budget $B$ limit the number of trials, each of which may involve multiple model calls or feedback steps. For $N$ failure episodes indexed by $i$, let $u_i$ and $s_{0,i}$ denote the request and initial state, and let $\pi_{i,t}$ be the final realization of trial $t$. With $y_{i,t}=y(\pi_{i,t};u_i,s_{0,i})$, the recovery rate is the fraction of failure episodes recovered within a budget of $B$ trials:
\begin{equation}
    \rec@B=\frac{1}{N}\sum_{i=1}^{N}\max_{1\leq t\leq B}y_{i,t}.
    \label{eq:recovery}
\end{equation}
Recovery@$B$ measures offline whether any of the $B$ trials succeeds. Let $a_i$ denote the upfront repair time for episode $i$ and $c_{i,t}$ the incremental time of trial $t$, including generation, execution, and any replay. The full-budget repair time and its mean over episodes are
\begin{equation}
    C_i^{\mathrm{full}}(B)=a_i+\sum_{t=1}^{B}c_{i,t},\qquad
    \overline{C}^{\mathrm{full}}(B)=\frac{1}{N}\sum_{i=1}^{N}C_i^{\mathrm{full}}(B).
    \label{eq:cost}
\end{equation}
We seek a favorable trade-off between recovery rate $\rec@B$ and mean full-budget repair time $\overline{C}^{\mathrm{full}}(B)$. The central design question is how to organize search across operation supports and within each support, reusing operation-level proposal computation while preserving flexibility in entity bindings, arguments, and action composition. Section~\ref{sec:method} develops ReCommit to realize this hierarchy through a shared readout from a dLLM for support ranking and support-conditioned realization search for complementary repairs.

%% file: sections/method.tex
\section{ReCommit}
\label{sec:method}
\begin{figure}[t]
    \centering
    \includegraphics[width=0.98\linewidth]{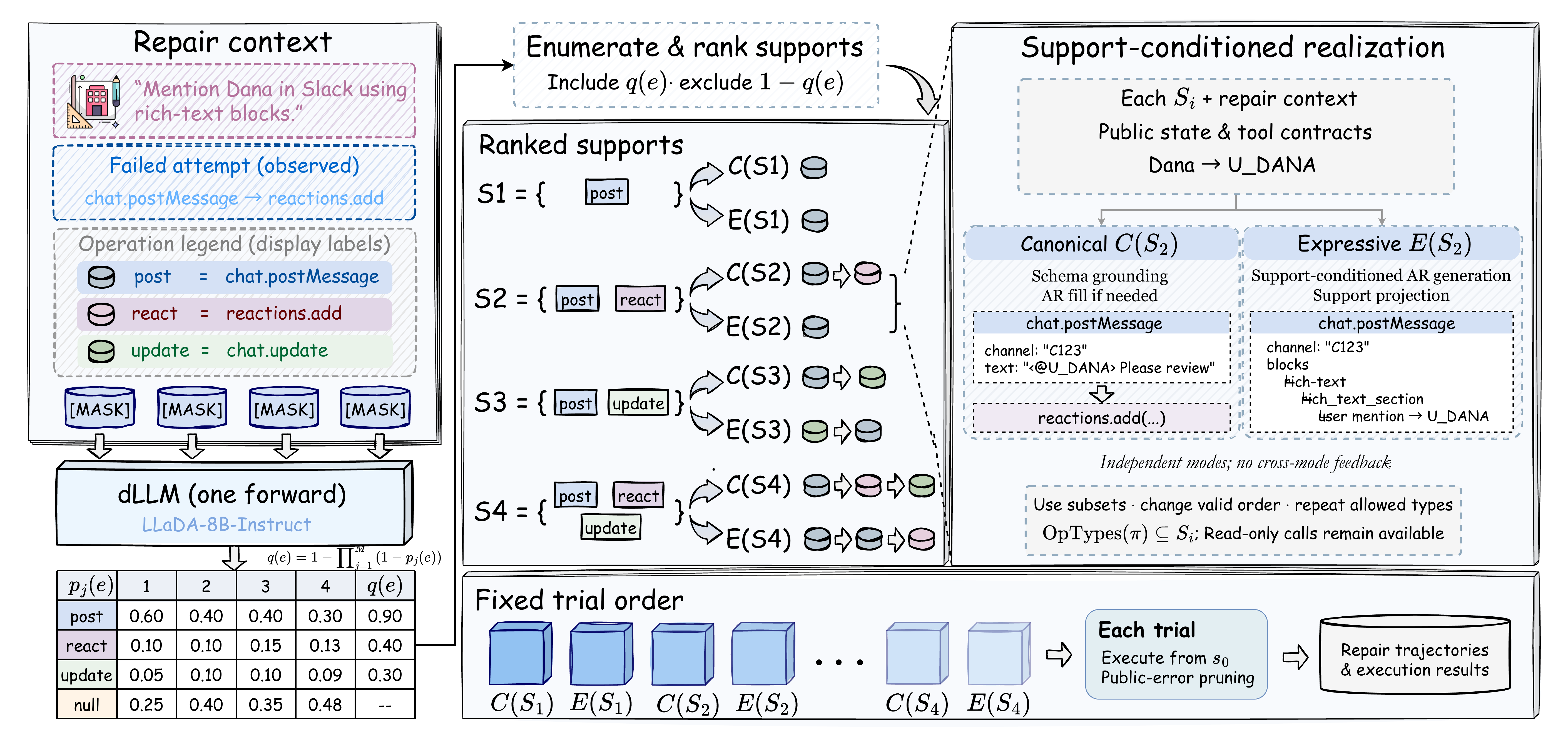}
    \vspace{-10pt}
    \caption{Overview of ReCommit. AR denotes autoregressive. Left: a single parallel dLLM readout supplies slot-level operation predictions, pooled into inclusion scores for support enumeration and ranking. Center: ranked supports define realization regions for different call sequences. Right: canonical realization $C(S_i;x)$ grounds a schema-based call structure, while expressive realization $E(S_i;x)$ searches alternative entity bindings, arguments, and action composition within the same support. Both modes use the original public context without cross-mode generation feedback. Bottom: repair trials follow the fixed order $C(S_1;x),E(S_1;x),C(S_2;x),E(S_2;x),\ldots$. Within each trial, public-error pruning removes rejected calls and replays the remaining nonempty sequence once from the initial state.}
    \label{fig:method}
    \vspace{-10pt}
\end{figure}

ReCommit turns a single diffusion readout into a reusable basis for hierarchical repair search. It realizes the formulation in Section~\ref{sec:problem} through \emph{amortized support search} and \emph{support-conditioned realization search}, with operation supports linking the two levels. Across supports, a shared parallel readout from a dLLM provides operation-type scores that guide search over alternative operation choices. Within each support, realization search respects the permitted operation types while varying entity bindings, arguments, and action composition. This separation allows search to vary both operation choices and their realizations without repeating the dLLM proposal computation at every trial. Public-error pruning then processes rejected calls within each trial. The framework operates on public repair information using pretrained models without additional training. Figure~\ref{fig:method} provides a detailed overview of the hierarchical repair process.

\subsection{Amortized Support Search}
\label{sec:support}
Amortized support search derives reusable operation-level guidance from the parallel masked predictions of a dLLM~\citep{llada,dream}. ReCommit uses one shared readout to rank alternative operation supports, leaving concrete tool-call sequences to realization search.

The dLLM receives the request, failed attempt, and an operation legend from the public context $x$, followed by $M$ masked slots. Each state-changing operation type is represented by a single-token code, with an additional null code for an unused slot. A single forward pass produces logits at all masked slots in parallel. ReCommit applies softmax over the operation and null codes at each slot to obtain distributions $p_j(e\mid x)$ for $j=1,\ldots,M$. These distributions form the shared diffusion readout used throughout support search.

The masked slots provide parallel predictions of operation types, whereas an operation support records which types are permitted regardless of slot position. We therefore pool the dLLM predictions into an inclusion score for each type $e\in\mathcal{T}$:
\begin{equation}
    q_e=1-\prod_{j=1}^{M}\bigl(1-p_j(e\mid x)\bigr).
    \label{eq:inclusion}
\end{equation}
Pooling aggregates evidence for including $e$ across slots and is invariant to slot permutations. It shifts search from slot assignments to operation-type membership, leaving call order and multiplicity to realization search. The inclusion scores define a factorized support ranking score:
\begin{equation}
    \ell(S)=\sum_{e\in S}\log q_e+\sum_{e\in\mathcal{T}\setminus S}\log(1-q_e).
    \label{eq:support}
\end{equation}
A highest-scoring support is $S_{\mathrm{base}}=\{e\in\mathcal{T}:q_e>1/2\}$. For any other support, the score decrease decomposes over the types whose membership changes:
\begin{equation}
    \ell(S_{\mathrm{base}})-\ell(S)
    =\sum_{e\in S\triangle S_{\mathrm{base}}}
    \left|\log\frac{q_e}{1-q_e}\right|,
    \label{eq:support-penalty}
\end{equation}
where $\triangle$ denotes symmetric difference. ReCommit enumerates distinct supports by increasing cumulative score decrease, subject to a bound on the number of membership changes. Types with inclusion scores near $1/2$ incur smaller penalties, so early alternatives reconsider less decisive operation choices. The resulting ranking $S_1,S_2,\ldots$ reuses the same dLLM scores to propose alternative operation choices, amortizing operation-level model computation across repair trials.

\subsection{Support-Conditioned Realization Search}
\label{sec:realizer}
For each support from the shared dLLM readout, realization search determines entity bindings, arguments, and action composition within $\mathcal{R}(S)$. ReCommit preserves the operation-level proposal while allowing the call structure to vary across realizations. The canonical mode grounds a schema-based call structure, while the expressive mode searches beyond it within the same operation support.

\noindent\textbf{Canonical realization.}
The canonical mode expands the permitted operation types in $S$ into a call structure using service-specific schema-based expansion rules and read dependencies, then grounds this structure to construct $C(S;x)$. Grounding resolves entities and task-specified values from the request and available public state. When public evidence suffices to fill the arguments, ReCommit adopts the grounded sequence directly. Otherwise, an autoregressive model fills unresolved arguments while preserving the grounded fields and call structure. This mode preserves choices established by public evidence and concentrates generation on the remaining arguments.

\noindent\textbf{Expressive realization.}
To search beyond the canonical call structure, the expressive mode constructs $E(S;x)$ by conditioning an autoregressive model on the permitted types, request, failed attempt, visible state, and tool contracts. It generates a complete call sequence that may use a subset of the permitted types, vary bindings and arguments, repeat operations, and change call order. ReCommit projects the generated sequence onto $\mathcal{R}(S)$ by removing state-changing calls outside $S$ while retaining known public read-only calls. Both modes construct their realizations independently from the original public context $x$. Together, they search alternative concrete realizations while reusing the same dLLM operation-level proposal.

Following the support ranking from the shared dLLM scores, ReCommit attempts a canonical realization followed by an expressive realization before moving to the next support: $C(S_1;x),E(S_1;x),C(S_2;x),E(S_2;x),\ldots$.
Trial $t$ uses support index $k_t=\lceil t/2\rceil$, so a budget of $B$ trials visits $\lceil B/2\rceil$ ranked supports. This fixed schedule first attempts schema-based grounding, then searches beyond the canonical call structure within the same support before considering another operation choice. It allocates trials to search both across supports and within each support.

\subsection{Execution with Public-Error Pruning}
A realization can contain useful operations alongside calls rejected by the environment. ReCommit applies \emph{public-error pruning} to replay the remaining calls within the same repair trial.

Let $\widetilde{\pi}_t$ denote the realization returned by the canonical or expressive mode scheduled at trial $t$. ReCommit executes it from $s_0$ and identifies rejected calls from public execution observations. If any calls are rejected and removing them leaves a nonempty sequence, the framework replays that sequence once from $s_0$ and directly adopts it as the final realization $\pi_t$. Otherwise, $\pi_t=\widetilde{\pi}_t$. The trial outcome is evaluated on the final execution. Pruning preserves the support constraint because removing calls cannot introduce a new operation type.

Under Eq.~\ref{eq:cost}, upfront time $a_i$ includes the shared dLLM readout and support enumeration, while incremental time $c_{i,t}$ includes realization, execution, and any pruning replay.

%% file: sections/experiments.tex
\section{Experiments}
\label{sec:experiments}
To assess whether the proposed diffusion-guided hierarchical search improves recovery with less computation, we conduct a series of experiments to examine the overall performance of ReCommit and the contributions of both search levels. Specifically, we address the following main questions:
\begin{enumerate}
\setlength{\itemsep}{0pt}
  \item[\textbf{Q1:}] How effective is ReCommit at recovering tool-agent failures across services and diffusion backbones under fixed repair budgets?
  \item[\textbf{Q2:}] Does ReCommit achieve a better recovery--cost trade-off across repair budgets?
  \item[\textbf{Q3:}] How does diffusion-guided support search compare with autoregressive support search?
  \item[\textbf{Q4:}] How do within-support realization search and public-error pruning contribute to recovery?
\end{enumerate}
Our code is available at \url{https://github.com/X-Xia0828/ReCommit}.

\begin{table}[t]
\centering
\caption{Fixed-budget recovery at $B=3$ and $B=13$. Service columns report recovered episodes from Box (129), Calendar (180), Linear (156), and Slack (166). Recovery rate is episode-weighted (\%), and time denotes mean full-budget repair time in seconds. Best and second-best values within each model-scale group at each budget are \textbf{bolded} and \underline{underlined}. ReCommit uses LLaDA-8B for proposal and Qwen3-8B for realization. Comparison methods use Qwen3-8B and Qwen3-32B.}

\label{tab:q1}
\small
\setlength{\tabcolsep}{1.5pt}
\renewcommand{\arraystretch}{1.0}
\resizebox{\linewidth}{!}{%
\begin{tabular}{@{}cccccccccccccc@{}}
\toprule
\multirow[c]{2}{*}{\textbf{Scale}} & \multirow[c]{2}{*}{\textbf{Method}} & \multicolumn{6}{c}{\textbf{Budget} $B=3$} & \multicolumn{6}{c}{\textbf{Budget} $B=13$} \\
\cmidrule(lr){3-8}\cmidrule(lr){9-14}
& & \textbf{Box} & \textbf{Calendar} & \textbf{Linear} & \textbf{Slack} & \textbf{Rate} $\uparrow$ & \textbf{Time} $\downarrow$ & \textbf{Box} & \textbf{Calendar} & \textbf{Linear} & \textbf{Slack} & \textbf{Rate} $\uparrow$ & \textbf{Time} $\downarrow$ \\
\midrule
\multirow[c]{3}{*}{\textbf{32B}} & Direct & 12 & 12 & 17 & 57 & 15.5 & \textbf{108.0} & 22 & 14 & 17 & 65 & 18.7 & \textbf{487.1} \\
 & Schema & \underline{30} & \textbf{23} & \underline{25} & \underline{65} & \underline{22.7} & \underline{143.1} & \underline{42} & \textbf{31} & \underline{27} & \underline{67} & \underline{26.5} & \underline{649.8} \\
 & Feedback-Reflective & \textbf{41} & \underline{20} & \textbf{32} & \textbf{70} & \textbf{25.8} & 350.4 & \textbf{56} & \underline{29} & \textbf{34} & \textbf{85} & \textbf{32.3} & 1528.3 \\
\midrule
\multirow[c]{9}{*}{\textbf{8B}} & Direct & 4 & 2 & 3 & 26 & 5.5 & \underline{40.8} & 8 & 2 & 3 & 34 & 7.4 & 177.7 \\
 & Schema & 10 & 7 & 18 & 31 & 10.5 & 44.8 & 13 & 9 & 18 & 36 & 12.0 & 197.0 \\
 & Reflective & 8 & 7 & 20 & 27 & 9.8 & 41.2 & 11 & 10 & 20 & 29 & 11.1 & \underline{177.6} \\
 & Feedback-Schema & 15 & 6 & 21 & 32 & 11.7 & 132.8 & 20 & 10 & 21 & \underline{37} & 13.9 & 576.9 \\
 & Feedback-Reflective & 15 & 7 & \underline{23} & 30 & 11.9 & 117.3 & 18 & 12 & \underline{23} & 33 & 13.6 & 517.1 \\
 & Plan-and-Execute & \underline{21} & \underline{9} & 13 & 31 & 11.7 & 62.0 & \underline{25} & 11 & 13 & 35 & 13.3 & 269.1 \\
 & ReAct-style & 15 & 5 & 13 & 19 & 8.2 & 84.0 & 17 & 5 & 14 & 26 & 9.8 & 361.2 \\
 & AgentDebug-style & 14 & \textbf{10} & 21 & \underline{34} & \underline{12.5} & 86.7 & 20 & \textbf{16} & 22 & \underline{37} & \underline{15.1} & 325.7 \\
\rowcolor{recommitTableHighlight}
 & \textcolor{recommitTableAccent}{\textbf{ReCommit}} & \textbf{23} & \underline{9} & \textbf{32} & \textbf{75} & \textbf{22.0} & \textbf{33.6} & \textbf{28} & \underline{13} & \textbf{34} & \textbf{80} & \textbf{24.6} & \textbf{158.7} \\
\bottomrule
\end{tabular}%
}
\end{table}

\subsection{Experimental Setup}
\noindent\textbf{Tasks.}
We construct a repair evaluation pool from the Agent-Diff benchmark~\citep{agentdiff}, covering Box, Calendar, Linear, and Slack. Source tool-use attempts are generated by Qwen3-8B~\citep{qwen3} using four random seeds, and only attempts that fail the benchmark task predicate enter the pool. The deduplicated pool contains 631 real failure episodes from 194 public tasks. All methods are evaluated on the same failure pool. Appendix~\ref{app:implementation} details pool construction and reports the task and episode counts for each service.

\noindent\textbf{Comparison methods.}
By default, ReCommit uses LLaDA-8B-Instruct~\citep{llada} as its dLLM for support proposal and Qwen3-8B~\citep{qwen3} for realization, with $M=8$ masked slots across all services. To evaluate generality across diffusion backbones, we also use Dream-v0-Instruct-7B~\citep{dream}, LLaDA-1.5-8B~\citep{llada15}, SDAR-8B~\citep{sdar}, DiffuCoder-7B~\citep{diffucoder}, and Nemotron-Diffusion-8B~\citep{nemotron}, keeping the realizer and search schedule fixed.

We compare against repair strategies using Qwen3-8B: Direct, Schema, Reflective~\citep{reflexion}, Feedback-Schema and Feedback-Reflective~\citep{critic}, Plan-and-Execute~\citep{planandsolve}, ReAct-style~\citep{react}, and AgentDebug-style~\citep{agentdebug}. All are adapted to the same tool-agent repair setting. We also evaluate Direct, Schema, and Feedback-Reflective with Qwen3-32B. Appendix~\ref{app:inference} details these methods and their implementations.

\noindent\textbf{Metrics.}
We measure recovery with Recovery@$B$ and computational cost with mean full-budget repair time, defined in Eqs.~\ref{eq:recovery}--\ref{eq:cost}. Q1 uses repair budgets $B=3$ and $B=13$, while Q2 evaluates $B\in\{1,3,5,7,9,11,13\}$. A successful repair requires both valid execution and satisfaction of the benchmark task assertions, following Eq.~\ref{eq:success}. Hidden assertion outcomes are used only for offline evaluation. Repair time is averaged over all failure episodes and includes proposal, realization, execution, and any pruning replay, excluding model loading. All experiments run on NVIDIA A100-SXM4-80GB GPUs. We report 95\% confidence intervals for recovery differences using paired bootstrap resampling at the task level (Appendix~\ref{app:measurement}).

\subsection{Main Experiments}
\noindent\textbf{To answer Q1.} Table~\ref{tab:q1} compares recovery across four services at repair budgets $B=3$ and $B=13$. ReCommit achieves higher aggregate recovery and lower mean full-budget repair time than all evaluated 8B comparison methods at both budgets. Even at $B=3$, it recovers 139 episodes, exceeding the strongest evaluated 8B comparison method at $B=13$, which recovers 95. \emph{Increasing the trial budget alone does not close the recovery gap for these comparison methods.}

ReCommit leads the evaluated 8B methods on Box, Linear, and Slack at both budgets. It matches the best 32B recovery on Linear at both budgets and exceeds all evaluated 32B methods on Slack at $B=3$. Calendar remains challenging across the evaluated methods, although recovery with ReCommit improves from 9 to 13 episodes as the budget increases from $B=3$ to $B=13$. Appendix~\ref{app:repair-cases} discusses the distinction between executable repairs and complete task fulfillment. In Appendix~\ref{app:backbones}, we replace the diffusion proposer with five alternative backbones while fixing the realizer and realization schedule. All retain higher aggregate recovery than the evaluated 8B comparison methods at $B=3$, although recovery varies across backbones. \emph{The framework's recovery advantage extends beyond the default LLaDA proposer.}

\begin{figure}[t]
    \centering
    \includegraphics[width=0.98\linewidth]{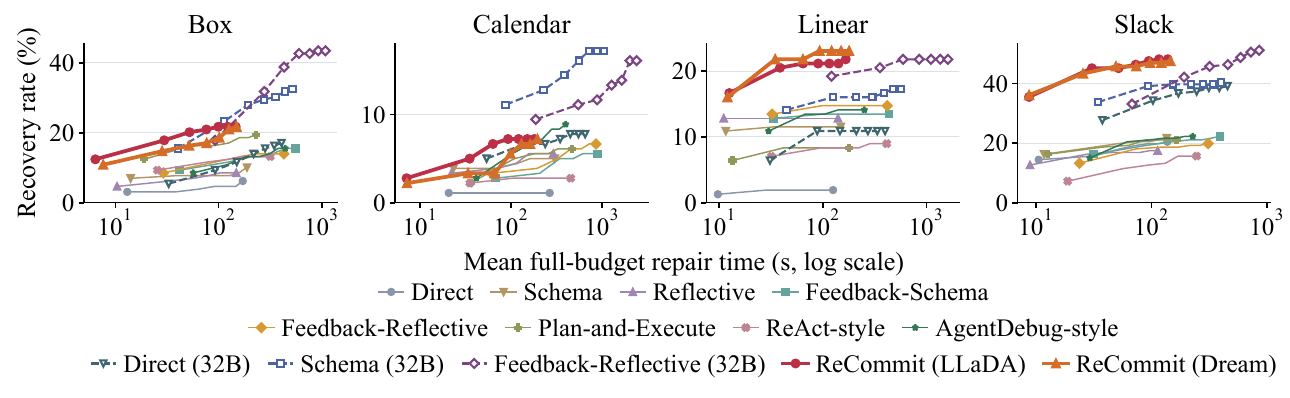}
\vspace{-10pt}
    \caption{Recovery--cost curves for four services with LLaDA- and Dream-based ReCommit and eleven comparison methods. Curves use $B\in\{1,3,5,7,9,11,13\}$. Only endpoints are marked for 8B comparison methods. Every point covers the complete service pool. Lines follow budget order on a logarithmic time axis to display the full cost range.}
    \label{fig:frontier}
\vspace{-15pt}
\end{figure}
\noindent\textbf{To answer Q2.} Figure~\ref{fig:frontier} compares recovery against full-budget repair time for each service. Increasing the budget continues to yield additional recoveries over the complete failure pool, but the incremental gains diminish as repair time grows. \emph{ReCommit achieves a favorable recovery--cost trade-off across the evaluated budgets.}

The 32B comparisons illustrate the cost of reaching similar recovery levels. At $B=5$, ReCommit achieves comparable recovery to Schema (32B) at $B=3$ (146 versus 143 episodes) with 59.3\% lower mean full-budget repair time. At $B=13$, Feedback-Reflective (32B) reaches 204 recoveries at greater cost. Proposal reuse avoids repeated operation scoring, while total repair time also depends on grounding, realization, and execution. Appendix~\ref{app:extensions} provides service-level recovery--cost results, paired comparisons, cost breakdowns, and offline first-success analysis, which further supports ReCommit's favorable recovery--cost trade-off.

\subsection{Mechanism Analysis}

\begin{figure}[t]
\centering
\includegraphics[width=0.90\linewidth]{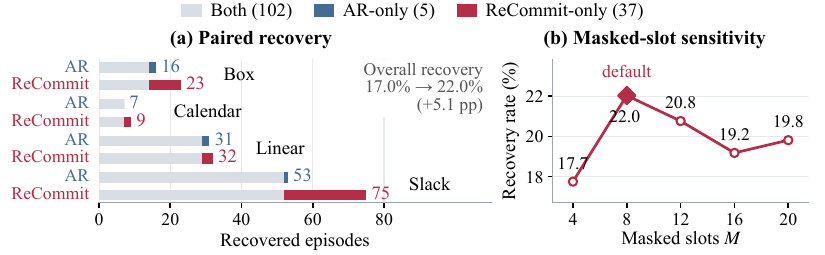}
\vspace{-8pt}
\caption{Support-proposer analysis at $B=3$ on 631 episodes. (a) Paired recovery against an autoregressive (AR) proposer. Gray segments denote episodes recovered by both configurations, colored segments exclusive recoveries, and end labels total recoveries. Both use the same realizer, pruning, and canonical/expressive schedule, with their own scoring and support ordering. (b) Aggregate recovery as the masked-slot count varies, with the shared default $M=8$ marked.}
\label{fig:q3paired}
\vspace{-10pt}
\end{figure}
\noindent\textbf{To answer Q3.} We compare ReCommit's diffusion-guided support proposer with an autoregressive (AR) control, varying operation scoring and support ordering while keeping the realizer, public-error pruning, and $C(S_1),E(S_1),C(S_2)$ schedule fixed. Appendix~\ref{app:arcontrol} details the two proposer configurations. Figure~\ref{fig:q3paired}(a) shows that both configurations recover a shared set of episodes, but ReCommit adds more exclusive recoveries, with the largest net gain on Slack. The aggregate recovery gain is 5.1 percentage points. \emph{With downstream realization and execution fixed, the evaluated diffusion-guided proposer improves repair coverage.}

Figure~\ref{fig:q3paired}(b) examines how the masked-slot count affects recovery. Aggregate recovery peaks at $M=8$ among the five evaluated settings; larger slot counts do not improve overall coverage. ReCommit uses the same default of $M=8$ across all services. Appendix~\ref{app:slots} provides the service-level results and timing measurements for all five evaluated settings.

\begingroup
\setlength{\intextsep}{0pt}
\begin{wraptable}{r}{0.50\textwidth}
\centering
\small
\setlength{\abovecaptionskip}{0pt}
\caption{Realization-search recovery. \emph{Supports} denotes the number of distinct supports visited. Without-P evaluates realizations without pruning. Trial budgets are equal, but computation differs. Best counts at each budget are \textbf{bolded}.}
\label{tab:q4}
\setlength{\tabcolsep}{2pt}
\renewcommand{\arraystretch}{1.0}
\resizebox{\linewidth}{!}{%
\begin{tabular}{@{}ccccc@{}}
\toprule
\multirow[c]{2}{*}{\makebox[16pt][c]{$B$}} & \multirow[c]{2}{*}{\textbf{Schedule}} & \multirow[c]{2}{*}{\textbf{Supports}} & \multicolumn{2}{c}{\textbf{Recovered episodes} $\uparrow$} \\
\cmidrule(lr){4-5}
& & & \textbf{With-P} & \textbf{Without-P} \\
\midrule
\multirow[c]{4}{*}{$3$} & Canonical-only & 3 & 127 & 117 \\
 & Expressive-only & 3 & 71 & 70 \\
 & Canonical resampling & 2 & 122 & 108 \\
\rowcolor{recommitTableHighlight}
 & \textcolor{recommitTableAccent}{\textbf{ReCommit}} & 2 & \textbf{139} & \textbf{128} \\
\midrule
\multirow[c]{4}{*}{$13$} & Canonical-only & 13 & 141 & 137 \\
 & Expressive-only & 13 & 85 & 83 \\
 & Canonical resampling & 7 & 139 & 136 \\
\rowcolor{recommitTableHighlight}
 & \textcolor{recommitTableAccent}{\textbf{ReCommit}} & 7 & \textbf{155} & \textbf{149} \\
\bottomrule
\end{tabular}
}
\end{wraptable}
\noindent\textbf{To answer Q4.} Table~\ref{tab:q4} compares four realization schedules with the same support proposer, support ranking, and trial budgets. ReCommit recovers more episodes than either single-mode schedule at both budgets, despite visiting fewer supports. Expressive-only trails canonical-only, yet alternating the two modes improves recovery, supporting their complementary coverage. To test whether ordinary repeated sampling offers similar gains, canonical resampling replaces each expressive trial with a second canonical sample on the same support. With the same trial and support counts, ReCommit recovers 17 and 16 additional episodes at $B=3$ and $B=13$, respectively. \emph{Within-support realization search improves recovery beyond both broader support coverage alone and the evaluated canonical-resampling policy.}

Public-error pruning raises recovery from 128 to 139 episodes at $B=3$ and from 149 to 155 at $B=13$. In Appendix~\ref{app:realization-search}, we report paired comparisons and analyze trial allocation. Appendices~\ref{app:execution} and~\ref{app:repair-cases} provide execution diagnostics and repair cases, respectively.
\par
\endgroup

%% file: sections/discussion.tex
\section{Conclusion and Discussion}
\label{sec:conclusion}
\noindent\textbf{Conclusion.} This work aims to improve tool-agent failure recovery while reducing the computational cost of repair. We formalize repair as hierarchical search over operation supports and their concrete realizations. ReCommit reuses a shared diffusion readout to guide support search while exploring complementary realizations within each support. Experiments show that ReCommit improves the recovery--cost trade-off in tool-agent repair.

\noindent\textbf{Discussion.} The fixed canonical/expressive schedule allocates trials to both search levels. However, fixed support rankings and trial allocation do not adapt to new execution evidence. The operation support abstraction allows grounding procedures to evolve without changing the shared operation-level proposal mechanism. In the future, we will investigate how to incorporate feedback-guided search allocation and transferable grounding into ReCommit to broaden repair coverage while preserving operation-level proposal reuse across repair trials.

%% file: sections/appendix.tex
\raggedbottom
\setlength{\textfloatsep}{12pt plus 2pt minus 2pt}
\setlength{\floatsep}{10pt plus 2pt minus 2pt}
\setlength{\intextsep}{10pt plus 2pt minus 2pt}
\makeatletter
\setlength{\@fptop}{0pt}
\setlength{\@fpsep}{8pt}
\setlength{\@fpbot}{0pt plus 1fil}
\makeatother

\section{Experimental Settings}
\label{app:implementation}

\begin{table}[!htb]
\centering
\renewcommand{\arraystretch}{1.13}
\small
\caption{Composition of the Agent-Diff repair evaluation pool. The task column counts distinct benchmark tasks; failure episodes are deduplicated unsuccessful tool-use attempts retained for repair. All methods are evaluated on the same pool.}
\label{tab:pool}
\begin{tabular}{@{}>{\raggedright\arraybackslash}p{\dimexpr0.328000\linewidth-1.333333\tabcolsep\relax}>{\raggedleft\arraybackslash}p{\dimexpr0.246000\linewidth-1.333333\tabcolsep\relax}>{\raggedleft\arraybackslash}p{\dimexpr0.246000\linewidth-1.333333\tabcolsep\relax}@{}}
\toprule
\rowcolor{recommitTableHeader}
\textbf{Service} & \textbf{Public tasks} & \textbf{Failure episodes} \\
\midrule
Box & 40 & 129 \\
Calendar & 53 & 180 \\
Linear & 50 & 156 \\
Slack & 51 & 166 \\
\rowcolor{recommitTableHighlight}
\textbf{Total} & 194 & 631 \\
\bottomrule
\end{tabular}
\end{table}
\noindent\textbf{Pool construction.}
We generate source tool-use attempts with Qwen3-8B~\citep{qwen3} using four random seeds. Attempts are deduplicated by public task and canonicalized proposed action, so equivalent attempts from different seeds contribute a single episode, while distinct failed attempts on the same task remain separate episodes. Only attempts that fail the benchmark task predicate enter the repair pool. Table~\ref{tab:pool} reports 631 failure episodes from 194 tasks, broken down by service.

\subsection{Methods and Implementation}
\label{app:inference}

\noindent\textbf{Enumeration.}
At each slot, logits over the operation and null codes are normalized by softmax. Null contributes no operation type to pooling, and operation probabilities are not renormalized after its removal. Retaining the null probability allows a slot to assign little mass to all operation types instead of redistributing that mass among them. We enumerate up to thirteen ranked supports per episode, with at most six membership changes relative to $S_{\mathrm{base}}$. This bound limits how many operation types are added or removed from the base support; it does not limit the number of calls in a realization. ReCommit uses the first seven supports at $B=13$, while canonical-only uses all thirteen. We use $M=8$ masked slots consistently across all four services. The slot count controls the parallel predictions pooled for operation scoring, whereas the repair budget controls how many realizations are attempted.

\noindent\textbf{Comparison methods.}
Direct generates a complete repair call list from the request, failed attempt, and public context. Schema adds public tool contracts to this input. Reflective uses the same schema-level input and adds a failure-reflection instruction inspired by Reflexion~\citep{reflexion}. Reflection is incorporated into the call-generation prompt.

Feedback-Schema and Feedback-Reflective adapt the tool-feedback refinement principle of CRITIC~\citep{critic}, conditioning revisions on preceding calls and public execution events. Feedback-Reflective also includes the reflection instruction.

Plan-and-Execute adapts the plan-then-solve decomposition of Plan-and-Solve~\citep{planandsolve}. It first generates an ordered list of tool names, then conditions a separate call-list generation on that plan. The ReAct-style strategy~\citep{react} generates one action at a time and receives public tool feedback before the next step. Each step produces a tool call or a finish response. AgentDebug-style~\citep{agentdebug} uses Qwen3-8B to generate repairs conditioned on trajectory diagnosis and critical-error feedback. It uses public execution evidence to update feedback between trials.

Direct (32B), Schema (32B), and Feedback-Reflective (32B) apply the corresponding repair strategies with Qwen3-32B. We adapt the cited repair strategies to the shared tool-agent repair setting. Public-error pruning is enabled for ReCommit and the corresponding ablations. The system comparison methods use their own execution procedures without this additional step.

\noindent\textbf{Models and decoding.}
The main checkpoints are \texttt{GSAI-ML/LLaDA-8B-Instruct}, \texttt{Qwen/Qwen3-8B}, and \texttt{Qwen/Qwen3-32B}. Models run in BF16 with Qwen thinking disabled. ReCommit uses greedy decoding; comparison methods use temperature 0.7, except that AgentDebug-style diagnosis and feedback updates use greedy decoding. Repair trials execute sequentially. Comparisons use the same repair-trial budgets; their computational costs are captured by mean full-budget repair time.

\noindent\textbf{Realization implementation.}
For example, a Linear support containing \texttt{issueUpdate} yields an update call preceded by public lookups required by its entity dependencies. Grounding can bind the label from the visible label catalog and the title from text explicitly quoted in the request. In canonical realization, values from the original failed attempt $\pi^{\mathrm{fail}}$ can initialize fields left unresolved by grounding. Qwen3-8B fills the remaining arguments while preserving grounded fields and the call structure. This separates the schema-based construction of calls from the generation of unresolved argument values. Expressive realization uses the original public context independently, without access to the canonical realization or its execution feedback. Expressive outputs containing unknown tools or malformed calls are treated as invalid realizations.

\noindent\textbf{Model identifiers.}
Table~\ref{tab:modelids} identifies the additional checkpoints. Adapters preserve each model's token and attention conventions, including SDAR's block-causal attention.
\begin{table}[!htb]
\centering
\renewcommand{\arraystretch}{1.13}\small
\caption{Diffusion backbones evaluated for support proposal beyond the default LLaDA-8B-Instruct. Repository identifiers specify the public Hugging Face checkpoints; all configurations use Qwen3-8B for realization under the same search schedule.}
\label{tab:modelids}
\begin{tabular}{@{}>{\raggedright\arraybackslash}p{\dimexpr0.230000\linewidth-1.000000\tabcolsep\relax}>{\raggedright\arraybackslash}p{\dimexpr0.770000\linewidth-1.000000\tabcolsep\relax}@{}}
\toprule
\rowcolor{recommitTableHeader}
\textbf{Backbone} & \textbf{Repository identifier} \\
\midrule
Dream & \path{Dream-org/Dream-v0-Instruct-7B} \\
LLaDA 1.5 & \path{GSAI-ML/LLaDA-1.5} \\
DiffuCoder & \path{apple/DiffuCoder-7B-Instruct} \\
SDAR & \path{JetLM/SDAR-8B-Chat-b8} \\
Nemotron & \path{nvidia/Nemotron-Labs-Diffusion-8B} \\
\bottomrule
\end{tabular}
\end{table}

\subsubsection{Operation Vocabularies}
\label{app:vocab}
Operation supports are subsets of the state-changing operation vocabularies in Table~\ref{tab:operation-vocabularies}, fixed across all tasks of each service. Read-only calls remain available during realization.

\begin{table}[!htb]
\centering
\renewcommand{\arraystretch}{1.13}
\small
\caption{State-changing operation types available to support search in each service. \emph{Types} gives the vocabulary size; each operation support is a subset of the listed types. Read-only calls remain available for public lookups during realization.}
\label{tab:operation-vocabularies}
\setlength{\tabcolsep}{5pt}
\begin{tabular}{@{}p{0.11\linewidth}p{0.07\linewidth}p{\dimexpr0.82\linewidth-4\tabcolsep\relax}@{}}
\toprule
\rowcolor{recommitTableHeader}
\textbf{Service} & \textbf{Types} & \textbf{Operation vocabulary} \\
\midrule
Box & 20 & \raggedright \texttt{DELETE}~\path{/collections/{id}}, \texttt{DELETE}~\path{/comments/{id}}, \texttt{DELETE}~\path{/files/{id}}, \texttt{DELETE}~\path{/folders/{id}}, \texttt{DELETE}~\path{/hubs/{id}}, \texttt{DELETE}~\path{/tasks/{id}}, \texttt{POST}~\path{/collections}, \texttt{POST}~\path{/comments}, \texttt{POST}~\path{/files/content}, \texttt{POST}~\path{/files/{id}/content}, \texttt{POST}~\path{/folders}, \texttt{POST}~\path{/hubs}, \texttt{POST}~\path{/hubs/{id}/manage_items}, \texttt{POST}~\path{/tasks}, \texttt{PUT}~\path{/collections/{id}}, \texttt{PUT}~\path{/comments/{id}}, \texttt{PUT}~\path{/files/{id}}, \texttt{PUT}~\path{/folders/{id}}, \texttt{PUT}~\path{/hubs/{id}}, \texttt{PUT}~\path{/tasks/{id}}. \tabularnewline
\midrule
Calendar & 25 & \raggedright \path{acl.delete}, \path{acl.insert}, \path{acl.patch}, \path{acl.update}, \path{acl.watch}, \path{calendarList.delete}, \path{calendarList.insert}, \path{calendarList.patch}, \path{calendarList.update}, \path{calendarList.watch}, \path{calendars.clear}, \path{calendars.delete}, \path{calendars.insert}, \path{calendars.patch}, \path{calendars.update}, \path{channels.stop}, \path{events.delete}, \path{events.import}, \path{events.insert}, \path{events.move}, \path{events.patch}, \path{events.quickAdd}, \path{events.update}, \path{events.watch}, \path{settings.watch}. \tabularnewline
\midrule
Linear & 13 & \raggedright \path{commentCreate}, \path{commentDelete}, \path{commentUpdate}, \path{issueCreate}, \path{issueLabelCreate}, \path{issueLabelDelete}, \path{issueLabelUpdate}, \path{issueRelationCreate}, \path{issueUpdate}, \path{teamCreate}, \path{teamMembershipCreate}, \path{workflowStateArchive}, \path{workflowStateCreate}. \tabularnewline
\midrule
Slack & 14 & \raggedright \path{chat.delete}, \path{chat.postMessage}, \path{chat.update}, \path{conversations.archive}, \path{conversations.create}, \path{conversations.invite}, \path{conversations.join}, \path{conversations.kick}, \path{conversations.leave}, \path{conversations.rename}, \path{conversations.setTopic}, \path{conversations.unarchive}, \path{reactions.add}, \path{reactions.remove}. \tabularnewline
\bottomrule
\end{tabular}
\end{table}

\subsection{Statistical Analysis}
\label{app:measurement}

We estimate confidence intervals for differences in recovery rates between methods. Each failure episode contributes a binary outcome indicating whether any repair trial succeeds within the specified budget. Multiple episodes can originate from the same benchmark task and share its request and environment, so we resample tasks while retaining their associated episodes as a group. Within each resampled pool, we compute each method's recovery rate over all sampled episodes, preserving the episode weighting used in Recovery@$B$. Tasks with more failure episodes contribute proportionally more to the recovery rate.

We use 10,000 paired bootstrap replicates. In each replicate, we draw the original number of tasks with replacement and include all episodes of each selected task, repeating the group when a task is sampled more than once. Both methods use the same sampled tasks and episodes, so the comparison remains paired. We compute the difference between their episode-weighted recovery rates and report the 2.5th and 97.5th percentiles of these differences as the 95\% confidence interval, expressed in percentage points. This procedure preserves the grouping of related failure episodes while aligning the reported uncertainty with the recovery differences evaluated in the main text.

\section{Realization Diagnostics and Cases}
\label{app:cases}

\subsection{Repair Cases}
\label{app:repair-cases}
The following cases illustrate support changes, realization differences, and incomplete repairs.

\noindent\textbf{Linear: changing the permitted operations.}
In Figure~\ref{fig:motivation}(a), the first support contains \texttt{issueLabelCreate} and \texttt{issueLabelUpdate}. These operations act on labels, but the request requires attaching the UX label to an existing issue. Both canonical and expressive realizations under this support fail. The next support adds \texttt{issueUpdate}, and its canonical realization succeeds. This comparison identifies the missing decision: revising how label operations are realized does not supply the issue-update operation needed to modify the target issue.

\noindent\textbf{Slack: changing the mention representation.}
In Figure~\ref{fig:motivation}(b), both realizations use \texttt{chat.postMessage} and resolve the intended user's ID. Canonical realization places that ID in a textual mention, whereas expressive realization includes it in a user element within rich-text blocks. Both calls execute, but only the structured representation fulfills the request. The difference is therefore the representation of an already resolved entity, rather than selecting a different operation or finding a different user.

\noindent\textbf{Calendar: execution validity without completion.}
A Calendar request combines calendar creation, access control, multiple events, an update, and cancellation handling. The repair attempts include executable event operations without completing the whole request. Among the 167 unrecovered Calendar episodes at $B=13$, each has at least one execution-valid repair trial. These cases illustrate the gap between execution-valid repairs and complete fulfillment of multi-part requests.

\subsection{Realization Search and Pruning}
\label{app:realization-search}

\noindent\textbf{Raw realization diversity.}
We count distinct call sequences per episode, preserving operation order, repeated calls, and argument values, and including invalid sequences. At $B=13$, canonical-only produces 12.34 distinct sequences on average versus ReCommit's 9.18, yet recovers fewer episodes. This count measures variation in generated sequences, including changes that do not lead to task completion. A method can produce many distinct unsuccessful realizations for an episode without increasing recovery. The comparison therefore supports evaluating diversity through the additional episodes recovered, alongside the number of distinct sequences generated.

\noindent\textbf{Realization complementarity.}
Expressive-only uses the first $B$ expressive realizations; recovery without pruning uses corresponding unpruned executions. These controls compare recovery under matched repair-trial budgets. On the full pool, 21 episodes have a successful $E(S_1)$ while both $C(S_1)$ and $C(S_2)$ fail. Eleven have a successful $C(S_2)$ after both first-support realizations fail. These outcomes show complementary recoveries from searching within a support and moving to another support. At $B=13$, ReCommit recovers 20 episodes missed by canonical-only and misses six recovered by that control. The task-level 95\% confidence interval for the recovery gain is $[0.16,4.51]$ percentage points relative to canonical-only.

\noindent\textbf{Alternative realization versus repeated sampling.}
The canonical-resampling control replaces $E(S_i)$ with a second canonical sample $C^{\prime}(S_i)$ using the same support and canonical prompt at temperature 0.7. The alternating $C/C^{\prime}$ schedule is truncated at budget $B$, matching ReCommit's trial counts and support allocation. With pruning, resampling recovers 122 and 139 episodes at $B=3$ and $B=13$, versus ReCommit's 139 and 155. ReCommit has 19 paired wins and two losses at $B=3$, and 18 wins and two losses at $B=13$; the task-level 95\% confidence intervals for the recovery gains are $[1.09,4.59]$ and $[0.77,4.69]$ percentage points. Without pruning, the $B=13$ comparison yields 19 wins and 6 losses, with an interval of $[0.15,4.33]$ points. These results support expressive realization over the evaluated canonical-resampling policy.

\noindent\textbf{Support coverage and realization allocation.}
At $B=13$, we evaluate the first $K$ canonical and first $D$ expressive realizations, where $K+D=13$ and $D\leq K$. Each of the first $D$ supports receives both a canonical and an expressive realization, while the remaining $K-D$ supports receive only a canonical realization. Increasing $D$ therefore allocates more trials to alternative realizations within visited supports and fewer to additional supports. Table~\ref{tab:allocation} shows that replacing the thirteenth canonical trial with one expressive trial raises recovery from 141 to 154 episodes. Allocations with expressive trials recover 154--157 episodes, with the default allocation recovering 155. The main gain comes from including within-support exploration; further reallocations within the fixed budget yield smaller differences in aggregate recovery.

\begin{table}[!htb]
\centering
\renewcommand{\arraystretch}{1.13}\small
\caption{Recovery under canonical/expressive trial allocations at $B=13$. Each row uses the first $K$ canonical and first $D$ expressive realizations, with $K+D=13$ and $D\leq K$. With-P and Without-P report recovered episodes with and without public-error pruning. Trial budgets are equal, but computation differs. The default $(K,D)=(7,6)$ is shaded; column maxima are bolded.}
\label{tab:allocation}
\begin{tabular}{@{}>{\raggedleft\arraybackslash}p{\dimexpr0.232200\linewidth-1.500000\tabcolsep\relax}>{\raggedleft\arraybackslash}p{\dimexpr0.232200\linewidth-1.500000\tabcolsep\relax}>{\raggedleft\arraybackslash}p{\dimexpr0.197800\linewidth-1.500000\tabcolsep\relax}>{\raggedleft\arraybackslash}p{\dimexpr0.197800\linewidth-1.500000\tabcolsep\relax}@{}}
\toprule
\rowcolor{recommitTableHeader}
\textbf{Canonical} $K$ & \textbf{Expressive} $D$ & \textbf{With-P} & \textbf{Without-P} \\
\midrule
13 & 0 & 141 & 137 \\
12 & 1 & 154 & 151 \\
11 & 2 & 156 & \textbf{153} \\
10 & 3 & 155 & 151 \\
9 & 4 & \textbf{157} & 152 \\
8 & 5 & 156 & 151 \\
\rowcolor{recommitTableHighlight}
7 & 6 & 155 & 149 \\
\bottomrule
\end{tabular}
\end{table}

\noindent\textbf{Action composition within a support.}
Among the 18 episodes recovered by ReCommit but not canonical resampling at $B=13$, twelve have a first successful expressive realization containing repeated state-changing operation types. Retaining only the first or only the last call of each type causes the same ten episodes, spanning six tasks, to fail the task assertions; nine retain valid execution. For example, \texttt{issueCreate} can be invoked three times to fulfill a request for three issues. These interventions show that repeated calls within one support are necessary to complete the multi-action requests in the ten inspected realizations.

\subsection{Execution Diagnostics}
\label{app:execution}
\noindent\textbf{Execution validity.}
Executable realization rate (ERR) is the fraction of repair trials whose final call sequence is nonempty and whose final execution has no rejected calls. We evaluate each method's final realization after its own generation and execution processing, including public-error pruning when enabled. Table~\ref{tab:err} compares all methods at $B=3$ on the same 631 episodes, giving 1,893 repair trials per method. ERR counts one final realization per repair trial. ERR weights each repair trial equally, whereas Recovery@$B$ counts an episode once if any trial completes the task. Several executable but unsuccessful trials can therefore increase ERR without improving episode-level recovery. Reporting both metrics distinguishes the ability to produce executable sequences from the ability to fulfill the original requests.

\begin{table}[!htb]
\centering
\renewcommand{\arraystretch}{1.13}
\small
\caption{Execution validity and task recovery at $B=3$. The executable-trial count includes nonempty final realizations with no rejected calls. Executable realization rate (ERR) divides this count by the 1,893 repair trials. Recovered episodes additionally require task completion. Realization controls use the same support proposer as ReCommit at this budget.}
\label{tab:err}
\setlength{\tabcolsep}{6pt}
\begin{tabular}{@{}>{\raggedright\arraybackslash}p{\dimexpr0.400000\linewidth-1.500000\tabcolsep\relax}>{\raggedleft\arraybackslash}p{\dimexpr0.220000\linewidth-1.500000\tabcolsep\relax}>{\raggedleft\arraybackslash}p{\dimexpr0.180000\linewidth-1.500000\tabcolsep\relax}>{\raggedleft\arraybackslash}p{\dimexpr0.200000\linewidth-1.500000\tabcolsep\relax}@{}}
\toprule
\rowcolor{recommitTableHeader}
\textbf{Method} & \textbf{Executable trials} & \textbf{ERR (\%)} & \textbf{Recovered episodes} \\
\midrule
\rowcolor{recommitTableHeader}
\multicolumn{4}{c}{\textbf{32B comparison methods}} \\
\midrule
Direct (32B) & 771 & 40.73 & 98 \\
Schema (32B) & 871 & 46.01 & 143 \\
Feedback-Reflective (32B) & 1,067 & 56.37 & 163 \\
\midrule
\rowcolor{recommitTableHeader}
\multicolumn{4}{c}{\textbf{8B methods}} \\
\midrule
Direct & 611 & 32.28 & 35 \\
Schema & 748 & 39.51 & 66 \\
Reflective & 825 & 43.58 & 62 \\
Feedback-Schema & 845 & 44.64 & 74 \\
Feedback-Reflective & 919 & 48.55 & 75 \\
Plan-and-Execute & 802 & 42.37 & 74 \\
ReAct-style & 992 & 52.40 & 52 \\
AgentDebug-style & 763 & 40.31 & 79 \\
\rowcolor{recommitTableHighlight}
\textcolor{recommitTableAccent}{\textbf{ReCommit}} & 1,694 & 89.49 & 139 \\
\midrule
\rowcolor{recommitTableHeader}
\multicolumn{4}{c}{\textbf{Realization controls}} \\
\midrule
Canonical-only & 1,832 & 96.78 & 127 \\
Expressive-only & 1,404 & 74.17 & 71 \\
\bottomrule
\end{tabular}
\end{table}

ReCommit achieves an ERR of 89.49\%, compared with 32.28--56.37\% across the eleven comparison methods. Canonical-only reaches 96.78\% but recovers 127 episodes, compared with ReCommit's 139. At $B=13$, canonical-only and ReCommit achieve ERR values of 96.66\% and 85.80\%, while recovering 141 and 155 episodes, respectively. Thus, higher execution validity alone does not imply broader recovery of the original user requests.

\section{Generality across Diffusion Backbones}
\label{app:backbones}

Table~\ref{tab:backbones} changes the diffusion backbone and adapter with the realizer and realization order fixed. Dream-v0-Instruct-7B~\citep{dream} (Dream-7B in the tables) recovers 131 episodes at $B=3$ in 30.67 seconds. Across the five alternative backbones, $B=3$ recovery ranges from 103 to 140 episodes, above the strongest evaluated 8B comparison method's 79. Linear recovery varies only from 32 to 34 episodes across backbones, whereas Box recovery ranges from 4 to 24. The backbone with the highest aggregate recovery does not lead every service: LLaDA-1.5 has the highest Box recovery, Dream and DiffuCoder tie for the highest Linear recovery, and the default LLaDA has the highest Slack recovery. This variation shows why the aggregate comparison is complemented by service-level results. These results support the use of different diffusion families within the same hierarchy, although recovery and timing vary across backbones and individual services.

\begin{table}[!htb]
\centering\small
\setlength{\tabcolsep}{3pt}
\renewcommand{\arraystretch}{1.13}
\caption{Diffusion-backbone comparison at $B=3$ with a fixed realizer and realization schedule. Service columns and Total count recovered episodes; Rate is episode-weighted recovery (\%), and Time is mean full-budget repair time (s). The default is shaded; column-best values are bolded.}
\label{tab:backbones}
\begin{tabular}{@{}>{\raggedright\arraybackslash}p{\dimexpr0.430000\linewidth-1.750000\tabcolsep\relax}>{\raggedleft\arraybackslash}p{\dimexpr0.060000\linewidth-1.750000\tabcolsep\relax}>{\raggedleft\arraybackslash}p{\dimexpr0.095000\linewidth-1.750000\tabcolsep\relax}>{\raggedleft\arraybackslash}p{\dimexpr0.075000\linewidth-1.750000\tabcolsep\relax}>{\raggedleft\arraybackslash}p{\dimexpr0.065000\linewidth-1.750000\tabcolsep\relax}>{\raggedleft\arraybackslash}p{\dimexpr0.075000\linewidth-1.750000\tabcolsep\relax}>{\raggedleft\arraybackslash}p{\dimexpr0.090000\linewidth-1.750000\tabcolsep\relax}>{\raggedleft\arraybackslash}p{\dimexpr0.110000\linewidth-1.750000\tabcolsep\relax}@{}}
\toprule
\rowcolor{recommitTableHeader}
\textbf{Proposal backbone} & \multicolumn{5}{c}{\textbf{Recovered episodes} $\uparrow$} & \textbf{Rate} & \textbf{Time} \\
\cmidrule(lr){2-6}
\rowcolor{recommitTableHeader}
& Box & Calendar & Linear & Slack & Total & (\%) $\uparrow$ & (s) $\downarrow$ \\
\midrule
\rowcolor{recommitTableHighlight}
LLaDA-8B & 23 & \textbf{9} & 32 & \textbf{75} & 139 & 22.03 & 33.60 \\
Dream-7B~\citep{dream} & 19 & 6 & \textbf{34} & 72 & 131 & 20.76 & 30.67 \\
LLaDA-1.5-8B~\citep{llada15} & \textbf{24} & \textbf{9} & 33 & 74 & \textbf{140} & \textbf{22.19} & \textbf{30.52} \\
SDAR-8B~\citep{sdar} & 19 & 5 & 33 & 69 & 126 & 19.97 & 31.93 \\
DiffuCoder-7B~\citep{diffucoder} & 15 & \textbf{9} & \textbf{34} & 66 & 124 & 19.65 & 34.99 \\
Nemotron-Diffusion-8B~\citep{nemotron} & 4 & 1 & 32 & 66 & 103 & 16.32 & 33.86 \\
\bottomrule
\end{tabular}
\end{table}

\section{Recovery--Cost Analysis}
\label{app:extensions}

\noindent\textbf{Where repair time is spent.}
Table~\ref{tab:costsplit} separates shared proposal time from downstream realization, execution, and pruning. Downstream work dominates total time: proposal reuse avoids repeated scoring, while end-to-end efficiency also depends on grounding and the generation performed by each strategy across its concrete repair trials.
\begin{table}[!htb]
\centering
\renewcommand{\arraystretch}{1.13}\small
\caption{ReCommit's mean full-budget repair time (s), split into shared operation scoring and support ranking, and downstream realization, execution, and pruning replay. Each budget includes the full proposal cost once for the complete repair search.}
\label{tab:costsplit}
\begin{tabular}{@{}>{\raggedleft\arraybackslash}p{\dimexpr0.086000\linewidth-1.500000\tabcolsep\relax}>{\raggedleft\arraybackslash}p{\dimexpr0.249400\linewidth-1.500000\tabcolsep\relax}>{\raggedleft\arraybackslash}p{\dimexpr0.292400\linewidth-1.500000\tabcolsep\relax}>{\raggedleft\arraybackslash}p{\dimexpr0.232200\linewidth-1.500000\tabcolsep\relax}@{}}
\toprule
\rowcolor{recommitTableHeader}
$B$ & \textbf{Shared proposal} & \textbf{Downstream work} & \textbf{Total} \\
\midrule
1 & 0.113 & 8.606 & 8.718 \\
3 & 0.113 & 33.486 & 33.599 \\
7 & 0.113 & 83.279 & 83.392 \\
13 & 0.113 & 158.556 & 158.669 \\
\bottomrule
\end{tabular}
\end{table}

Tables~\ref{tab:q2}--\ref{tab:q2-slack} report recovery and repair time for every comparison method and both ReCommit backbones on each service across all evaluated budgets. In aggregate, Feedback-Reflective (32B) has the highest recovery at $B=13$, with 204 episodes (32.33\%) and a mean time of 1,528.28 seconds. LLaDA- and Dream-based ReCommit recover 155 and 156 episodes in 158.67 and 168.69 seconds, respectively. These results illustrate similar recovery with different costs across the two diffusion backbones under the same repair-trial budget.

\noindent\textbf{Paired recovery comparisons.}
At $B=3$, ReCommit exceeds AgentDebug-style by 9.51 percentage points, with a paired task-level 95\% confidence interval of $[5.34,13.87]$. At $B=13$, the recovery gain is 9.51 percentage points, with an interval of $[5.12,14.11]$. For ReCommit at $B=5$ versus Schema (32B) at $B=3$, the recovery difference is 0.48 percentage points, with a 95\% confidence interval of $[-4.13,5.17]$ percentage points. The paired analysis finds no statistically significant recovery difference, while ReCommit reduces mean full-budget repair time by 59.3\%.

\noindent\textbf{Full-budget and first-success time.}
We additionally report \emph{offline first-success repair time}, which measures how early a correct repair appears within a fixed trial budget. Let $J_i(B)$ be the first successful trial within budget $B$, or $B$ if all trials fail. We define
\begin{equation}
    C_i^{\mathrm{first}}(B)=a_i+\sum_{t=1}^{J_i(B)}c_{i,t}.
    \label{eq:first-cost}
\end{equation}
This metric includes the complete upfront work and is averaged over all episodes, charging the full budget to failures. We compute this diagnostic offline using benchmark task assertions to identify the first successful trial. Tables~\ref{tab:first-success} and~\ref{tab:first-success-b13} compare all eleven baseline methods and both ReCommit variants at $B=3$ and $B=13$, respectively. The service-level means show how first-success time varies across task pools, while the aggregate recovery counts provide the quality context for each timing comparison.

With the LLaDA proposer, ReCommit achieves higher aggregate recovery and lower mean offline first-success time than every evaluated 8B comparison method at both budgets. Relative to AgentDebug-style, the strongest evaluated 8B comparison method by recovery, first-success time is reduced by 63.9\% at $B=3$ and 55.6\% at $B=13$. The recovery--cost advantage therefore also holds under the offline first-success metric.

\begin{table}[!htb]
\centering\small
\setlength{\tabcolsep}{3pt}
\renewcommand{\arraystretch}{1.13}
\caption{Offline first-success repair time at $B=3$. Service columns report mean time (s). Aggregate columns give recovered episodes and mean full-budget (Full) and offline first-success (First) times (s). All means include unrecovered episodes at their full-budget cost. Comparison methods use Qwen3-8B unless marked 32B; both ReCommit variants use Qwen3-8B for realization.}
\label{tab:first-success}
\begin{tabular}{@{}>{\raggedright\arraybackslash}p{\dimexpr0.290\linewidth-1.75\tabcolsep\relax}>{\raggedleft\arraybackslash}p{\dimexpr0.085\linewidth-1.75\tabcolsep\relax}>{\raggedleft\arraybackslash}p{\dimexpr0.100\linewidth-1.75\tabcolsep\relax}>{\raggedleft\arraybackslash}p{\dimexpr0.085\linewidth-1.75\tabcolsep\relax}>{\raggedleft\arraybackslash}p{\dimexpr0.085\linewidth-1.75\tabcolsep\relax}>{\raggedleft\arraybackslash}p{\dimexpr0.115\linewidth-1.75\tabcolsep\relax}>{\raggedleft\arraybackslash}p{\dimexpr0.110\linewidth-1.75\tabcolsep\relax}>{\raggedleft\arraybackslash}p{\dimexpr0.130\linewidth-1.75\tabcolsep\relax}@{}}
\toprule
\multirow[c]{2}{*}{\textbf{Method}} & \multicolumn{4}{c}{\textbf{First-success time by service}} & \multicolumn{3}{c}{\textbf{Aggregate}} \\
\cmidrule(lr){2-5}\cmidrule(lr){6-8}
 & Box & Calendar & Linear & Slack & Recovered & Full & First \\
\midrule
Direct (32B) & 89.3 & 138.6 & 86.0 & 91.3 & 98 & 108.0 & 103.1 \\
Schema (32B) & 102.1 & 213.1 & 118.8 & 77.3 & 143 & 143.1 & 131.4 \\
\shortstack[l]{Feedback-Reflective\\(32B)} & 237.0 & 516.1 & 332.6 & 160.8 & 163 & 350.4 & 320.2 \\
\midrule
Direct & 38.1 & 61.5 & 28.5 & 30.1 & 35 & 40.8 & 40.3 \\
Schema & 39.3 & 68.2 & 33.1 & 30.2 & 66 & 44.8 & 43.6 \\
Reflective & 33.2 & 65.9 & 31.2 & 26.0 & 62 & 41.2 & 40.1 \\
Feedback-Schema & 120.5 & 203.2 & 96.2 & 85.3 & 74 & 132.8 & 128.9 \\
Feedback-Reflective & 88.0 & 188.8 & 93.6 & 69.3 & 75 & 117.3 & 113.2 \\
Plan-and-Execute & 51.2 & 107.5 & 40.7 & 35.0 & 74 & 62.0 & 60.4 \\
ReAct-style & 71.6 & 102.4 & 91.0 & 53.5 & 52 & 84.0 & 80.4 \\
AgentDebug-style & 117.0 & 101.4 & 64.5 & 57.4 & 79 & 86.7 & 83.9 \\
\midrule
\rowcolor{recommitTableHighlight}
ReCommit (LLaDA) & 26.8 & 34.2 & 36.2 & 23.2 & 139 & 33.6 & 30.3 \\
\rowcolor{recommitTableHighlight}
ReCommit (Dream) & 26.9 & 33.0 & 32.5 & 19.7 & 131 & 30.7 & 28.1 \\
\bottomrule
\end{tabular}
\end{table}

\begin{table}[!htb]
\centering\small
\setlength{\tabcolsep}{3pt}
\renewcommand{\arraystretch}{1.13}
\caption{Offline first-success repair time at $B=13$. Service columns report mean time (s). Aggregate columns give recovered episodes and mean full-budget (Full) and offline first-success (First) times (s). All means include unrecovered episodes at their full-budget cost. Comparison methods use Qwen3-8B unless marked 32B; both ReCommit variants use Qwen3-8B for realization.}
\label{tab:first-success-b13}
\begin{tabular}{@{}>{\raggedright\arraybackslash}p{\dimexpr0.290\linewidth-1.75\tabcolsep\relax}>{\raggedleft\arraybackslash}p{\dimexpr0.085\linewidth-1.75\tabcolsep\relax}>{\raggedleft\arraybackslash}p{\dimexpr0.100\linewidth-1.75\tabcolsep\relax}>{\raggedleft\arraybackslash}p{\dimexpr0.085\linewidth-1.75\tabcolsep\relax}>{\raggedleft\arraybackslash}p{\dimexpr0.085\linewidth-1.75\tabcolsep\relax}>{\raggedleft\arraybackslash}p{\dimexpr0.115\linewidth-1.75\tabcolsep\relax}>{\raggedleft\arraybackslash}p{\dimexpr0.110\linewidth-1.75\tabcolsep\relax}>{\raggedleft\arraybackslash}p{\dimexpr0.130\linewidth-1.75\tabcolsep\relax}@{}}
\toprule
\multirow[c]{2}{*}{\textbf{Method}} & \multicolumn{4}{c}{\textbf{First-success time by service}} & \multicolumn{3}{c}{\textbf{Aggregate}} \\
\cmidrule(lr){2-5}\cmidrule(lr){6-8}
 & Box & Calendar & Linear & Slack & Recovered & Full & First \\
\midrule
Direct (32B) & 373.9 & 625.1 & 380.6 & 353.7 & 118 & 487.1 & 441.9 \\
Schema (32B) & 398.0 & 928.9 & 519.9 & 300.5 & 167 & 649.8 & 553.9 \\
\shortstack[l]{Feedback-Reflective\\(32B)} & 729.2 & 2154.3 & 1446.5 & 578.2 & 204 & 1528.3 & 1273.3 \\
\midrule
Direct & 168.9 & 264.2 & 124.9 & 124.8 & 47 & 177.7 & 173.6 \\
Schema & 181.0 & 293.3 & 141.2 & 121.0 & 76 & 197.0 & 187.4 \\
Reflective & 143.3 & 282.0 & 131.4 & 104.1 & 70 & 177.6 & 169.6 \\
Feedback-Schema & 515.8 & 866.8 & 408.7 & 344.0 & 88 & 576.9 & 544.3 \\
Feedback-Reflective & 387.4 & 827.1 & 390.6 & 275.1 & 86 & 517.1 & 484.1 \\
Plan-and-Execute & 205.2 & 451.7 & 173.5 & 141.9 & 84 & 269.1 & 251.0 \\
ReAct-style & 277.0 & 443.9 & 380.3 & 216.7 & 62 & 361.2 & 334.3 \\
AgentDebug-style & 411.4 & 380.0 & 232.3 & 194.1 & 95 & 325.7 & 301.0 \\
\midrule
\rowcolor{recommitTableHighlight}
ReCommit (LLaDA) & 124.3 & 168.8 & 148.1 & 89.8 & 155 & 158.7 & 133.8 \\
\rowcolor{recommitTableHighlight}
ReCommit (Dream) & 129.9 & 185.8 & 159.3 & 94.4 & 156 & 168.7 & 143.8 \\
\bottomrule
\end{tabular}
\end{table}

\begin{table}[!htbp]
\centering\small
\setlength{\tabcolsep}{4pt}
\renewcommand{\arraystretch}{1.18}
\caption{Recovery--cost results across budgets on Box (129 episodes). Cells give recovered episodes (top) and mean full-budget repair time in seconds (bottom). Comparison methods use Qwen3-8B unless marked 32B; both shaded ReCommit variants use Qwen3-8B for realization.}
\label{tab:q2}
\begin{tabular}{@{}>{\raggedright\arraybackslash}m{\dimexpr0.30\linewidth-1.75\tabcolsep\relax}>{\raggedleft\arraybackslash}m{\dimexpr0.10\linewidth-1.75\tabcolsep\relax}>{\raggedleft\arraybackslash}m{\dimexpr0.10\linewidth-1.75\tabcolsep\relax}>{\raggedleft\arraybackslash}m{\dimexpr0.10\linewidth-1.75\tabcolsep\relax}>{\raggedleft\arraybackslash}m{\dimexpr0.10\linewidth-1.75\tabcolsep\relax}>{\raggedleft\arraybackslash}m{\dimexpr0.10\linewidth-1.75\tabcolsep\relax}>{\raggedleft\arraybackslash}m{\dimexpr0.10\linewidth-1.75\tabcolsep\relax}>{\raggedleft\arraybackslash}m{\dimexpr0.10\linewidth-1.75\tabcolsep\relax}@{}}
\toprule
\rowcolor{recommitTableHeader}
\textbf{Method} & $B=1$ & $B=3$ & $B=5$ & $B=7$ & $B=9$ & $B=11$ & $B=13$ \\
\midrule
Direct (32B) & \shortstack[r]{7\\32.9} & \shortstack[r]{12\\92.7} & \shortstack[r]{15\\150.2} & \shortstack[r]{18\\218.9} & \shortstack[r]{20\\282.7} & \shortstack[r]{21\\346.5} & \shortstack[r]{22\\408.0} \\[2pt]
Schema (32B) & \shortstack[r]{20\\40.5} & \shortstack[r]{30\\113.5} & \shortstack[r]{36\\192.0} & \shortstack[r]{38\\274.7} & \shortstack[r]{39\\355.9} & \shortstack[r]{41\\439.2} & \shortstack[r]{42\\520.5} \\[2pt]
Feedback-Reflective (32B) & \shortstack[r]{23\\92.5} & \shortstack[r]{41\\277.7} & \shortstack[r]{50\\431.3} & \shortstack[r]{55\\602.9} & \shortstack[r]{55\\763.9} & \shortstack[r]{56\\924.0} & \shortstack[r]{56\\1084.8} \\[2pt]
\midrule
Direct & \shortstack[r]{4\\13.1} & \shortstack[r]{4\\38.3} & \shortstack[r]{5\\64.2} & \shortstack[r]{6\\90.6} & \shortstack[r]{6\\119.9} & \shortstack[r]{6\\146.8} & \shortstack[r]{8\\171.6} \\[2pt]
Schema & \shortstack[r]{9\\14.1} & \shortstack[r]{10\\40.3} & \shortstack[r]{10\\69.6} & \shortstack[r]{10\\98.6} & \shortstack[r]{10\\128.1} & \shortstack[r]{11\\157.4} & \shortstack[r]{13\\188.7} \\[2pt]
Reflective & \shortstack[r]{6\\10.4} & \shortstack[r]{8\\33.7} & \shortstack[r]{9\\57.1} & \shortstack[r]{10\\80.5} & \shortstack[r]{11\\103.6} & \shortstack[r]{11\\126.2} & \shortstack[r]{11\\148.3} \\[2pt]
Feedback-Schema & \shortstack[r]{12\\41.8} & \shortstack[r]{15\\125.1} & \shortstack[r]{17\\209.3} & \shortstack[r]{17\\293.4} & \shortstack[r]{19\\382.3} & \shortstack[r]{20\\469.5} & \shortstack[r]{20\\555.0} \\[2pt]
Feedback-Reflective & \shortstack[r]{11\\29.3} & \shortstack[r]{15\\92.8} & \shortstack[r]{17\\163.3} & \shortstack[r]{18\\230.0} & \shortstack[r]{18\\296.9} & \shortstack[r]{18\\362.8} & \shortstack[r]{18\\429.5} \\[2pt]
Plan-and-Execute & \shortstack[r]{16\\18.9} & \shortstack[r]{21\\53.7} & \shortstack[r]{21\\88.7} & \shortstack[r]{22\\125.3} & \shortstack[r]{24\\159.4} & \shortstack[r]{24\\194.9} & \shortstack[r]{25\\231.0} \\[2pt]
ReAct-style & \shortstack[r]{12\\25.5} & \shortstack[r]{15\\77.7} & \shortstack[r]{16\\124.5} & \shortstack[r]{17\\172.2} & \shortstack[r]{17\\220.5} & \shortstack[r]{17\\269.3} & \shortstack[r]{17\\317.7} \\[2pt]
AgentDebug-style & \shortstack[r]{11\\56.9} & \shortstack[r]{14\\120.4} & \shortstack[r]{17\\184.8} & \shortstack[r]{17\\249.1} & \shortstack[r]{18\\311.6} & \shortstack[r]{20\\376.5} & \shortstack[r]{20\\442.6} \\[2pt]
\midrule
\rowcolor{recommitTableHighlight}
ReCommit (LLaDA) & \shortstack[r]{16\\6.4} & \shortstack[r]{23\\29.7} & \shortstack[r]{26\\52.3} & \shortstack[r]{27\\76.2} & \shortstack[r]{28\\99.6} & \shortstack[r]{28\\122.5} & \shortstack[r]{28\\146.7} \\[2pt]
\rowcolor{recommitTableHighlight}
ReCommit (Dream) & \shortstack[r]{14\\7.6} & \shortstack[r]{19\\28.4} & \shortstack[r]{21\\51.9} & \shortstack[r]{22\\76.7} & \shortstack[r]{24\\101.7} & \shortstack[r]{27\\126.3} & \shortstack[r]{28\\149.4} \\[2pt]
\bottomrule
\end{tabular}
\end{table}

\begin{table}[!htbp]
\centering\small
\setlength{\tabcolsep}{4pt}
\renewcommand{\arraystretch}{1.18}
\caption{Recovery--cost results across budgets on Calendar (180 episodes). Cells give recovered episodes (top) and mean full-budget repair time in seconds (bottom). Comparison methods use Qwen3-8B unless marked 32B; both shaded ReCommit variants use Qwen3-8B for realization.}
\label{tab:q2-calendar}
\begin{tabular}{@{}>{\raggedright\arraybackslash}m{\dimexpr0.30\linewidth-1.75\tabcolsep\relax}>{\raggedleft\arraybackslash}m{\dimexpr0.10\linewidth-1.75\tabcolsep\relax}>{\raggedleft\arraybackslash}m{\dimexpr0.10\linewidth-1.75\tabcolsep\relax}>{\raggedleft\arraybackslash}m{\dimexpr0.10\linewidth-1.75\tabcolsep\relax}>{\raggedleft\arraybackslash}m{\dimexpr0.10\linewidth-1.75\tabcolsep\relax}>{\raggedleft\arraybackslash}m{\dimexpr0.10\linewidth-1.75\tabcolsep\relax}>{\raggedleft\arraybackslash}m{\dimexpr0.10\linewidth-1.75\tabcolsep\relax}>{\raggedleft\arraybackslash}m{\dimexpr0.10\linewidth-1.75\tabcolsep\relax}@{}}
\toprule
\rowcolor{recommitTableHeader}
\textbf{Method} & $B=1$ & $B=3$ & $B=5$ & $B=7$ & $B=9$ & $B=11$ & $B=13$ \\
\midrule
Direct (32B) & \shortstack[r]{9\\53.6} & \shortstack[r]{12\\141.0} & \shortstack[r]{12\\236.7} & \shortstack[r]{13\\345.5} & \shortstack[r]{14\\447.0} & \shortstack[r]{14\\549.6} & \shortstack[r]{14\\650.3} \\[2pt]
Schema (32B) & \shortstack[r]{20\\86.5} & \shortstack[r]{23\\226.6} & \shortstack[r]{26\\384.9} & \shortstack[r]{29\\553.8} & \shortstack[r]{31\\712.4} & \shortstack[r]{31\\880.4} & \shortstack[r]{31\\1048.3} \\[2pt]
Feedback-Reflective (32B) & \shortstack[r]{17\\186.1} & \shortstack[r]{20\\545.0} & \shortstack[r]{21\\881.3} & \shortstack[r]{24\\1271.1} & \shortstack[r]{25\\1643.4} & \shortstack[r]{29\\2022.7} & \shortstack[r]{29\\2398.2} \\[2pt]
\midrule
Direct & \shortstack[r]{2\\20.6} & \shortstack[r]{2\\61.7} & \shortstack[r]{2\\102.3} & \shortstack[r]{2\\143.3} & \shortstack[r]{2\\183.7} & \shortstack[r]{2\\224.4} & \shortstack[r]{2\\265.0} \\[2pt]
Schema & \shortstack[r]{7\\23.2} & \shortstack[r]{7\\69.3} & \shortstack[r]{8\\115.8} & \shortstack[r]{9\\161.5} & \shortstack[r]{9\\208.2} & \shortstack[r]{9\\254.9} & \shortstack[r]{9\\301.4} \\[2pt]
Reflective & \shortstack[r]{6\\22.0} & \shortstack[r]{7\\66.8} & \shortstack[r]{8\\112.5} & \shortstack[r]{10\\156.0} & \shortstack[r]{10\\201.1} & \shortstack[r]{10\\246.3} & \shortstack[r]{10\\290.9} \\[2pt]
Feedback-Schema & \shortstack[r]{5\\67.5} & \shortstack[r]{6\\205.5} & \shortstack[r]{9\\341.7} & \shortstack[r]{9\\480.0} & \shortstack[r]{10\\614.6} & \shortstack[r]{10\\751.8} & \shortstack[r]{10\\889.4} \\[2pt]
Feedback-Reflective & \shortstack[r]{6\\60.8} & \shortstack[r]{7\\191.9} & \shortstack[r]{9\\324.9} & \shortstack[r]{11\\455.8} & \shortstack[r]{11\\590.5} & \shortstack[r]{12\\723.7} & \shortstack[r]{12\\856.8} \\[2pt]
Plan-and-Execute & \shortstack[r]{4\\35.6} & \shortstack[r]{9\\108.3} & \shortstack[r]{10\\180.7} & \shortstack[r]{10\\255.7} & \shortstack[r]{11\\326.4} & \shortstack[r]{11\\397.6} & \shortstack[r]{11\\470.3} \\[2pt]
ReAct-style & \shortstack[r]{4\\35.0} & \shortstack[r]{5\\103.7} & \shortstack[r]{5\\174.0} & \shortstack[r]{5\\242.6} & \shortstack[r]{5\\312.6} & \shortstack[r]{5\\383.3} & \shortstack[r]{5\\453.0} \\[2pt]
AgentDebug-style & \shortstack[r]{5\\41.3} & \shortstack[r]{10\\102.7} & \shortstack[r]{12\\161.7} & \shortstack[r]{13\\219.9} & \shortstack[r]{15\\279.0} & \shortstack[r]{15\\337.3} & \shortstack[r]{16\\397.2} \\[2pt]
\midrule
\rowcolor{recommitTableHighlight}
ReCommit (LLaDA) & \shortstack[r]{5\\7.1} & \shortstack[r]{9\\34.9} & \shortstack[r]{12\\63.0} & \shortstack[r]{13\\92.3} & \shortstack[r]{13\\120.6} & \shortstack[r]{13\\150.2} & \shortstack[r]{13\\179.7} \\[2pt]
\rowcolor{recommitTableHighlight}
ReCommit (Dream) & \shortstack[r]{4\\7.2} & \shortstack[r]{6\\33.6} & \shortstack[r]{6\\65.7} & \shortstack[r]{10\\97.9} & \shortstack[r]{12\\129.8} & \shortstack[r]{12\\161.6} & \shortstack[r]{13\\194.6} \\[2pt]
\bottomrule
\end{tabular}
\end{table}

\begin{table}[!htbp]
\centering\small
\setlength{\tabcolsep}{4pt}
\renewcommand{\arraystretch}{1.18}
\caption{Recovery--cost results across budgets on Linear (156 episodes). Cells give recovered episodes (top) and mean full-budget repair time in seconds (bottom). Comparison methods use Qwen3-8B unless marked 32B; both shaded ReCommit variants use Qwen3-8B for realization.}
\label{tab:q2-linear}
\begin{tabular}{@{}>{\raggedright\arraybackslash}m{\dimexpr0.30\linewidth-1.75\tabcolsep\relax}>{\raggedleft\arraybackslash}m{\dimexpr0.10\linewidth-1.75\tabcolsep\relax}>{\raggedleft\arraybackslash}m{\dimexpr0.10\linewidth-1.75\tabcolsep\relax}>{\raggedleft\arraybackslash}m{\dimexpr0.10\linewidth-1.75\tabcolsep\relax}>{\raggedleft\arraybackslash}m{\dimexpr0.10\linewidth-1.75\tabcolsep\relax}>{\raggedleft\arraybackslash}m{\dimexpr0.10\linewidth-1.75\tabcolsep\relax}>{\raggedleft\arraybackslash}m{\dimexpr0.10\linewidth-1.75\tabcolsep\relax}>{\raggedleft\arraybackslash}m{\dimexpr0.10\linewidth-1.75\tabcolsep\relax}@{}}
\toprule
\rowcolor{recommitTableHeader}
\textbf{Method} & $B=1$ & $B=3$ & $B=5$ & $B=7$ & $B=9$ & $B=11$ & $B=13$ \\
\midrule
Direct (32B) & \shortstack[r]{10\\31.0} & \shortstack[r]{17\\88.4} & \shortstack[r]{17\\146.6} & \shortstack[r]{17\\210.6} & \shortstack[r]{17\\272.8} & \shortstack[r]{17\\336.0} & \shortstack[r]{17\\399.2} \\[2pt]
Schema (32B) & \shortstack[r]{22\\44.6} & \shortstack[r]{25\\124.8} & \shortstack[r]{25\\209.9} & \shortstack[r]{25\\301.0} & \shortstack[r]{26\\387.3} & \shortstack[r]{27\\476.5} & \shortstack[r]{27\\563.6} \\[2pt]
Feedback-Reflective (32B) & \shortstack[r]{30\\121.5} & \shortstack[r]{32\\355.9} & \shortstack[r]{34\\594.5} & \shortstack[r]{34\\851.3} & \shortstack[r]{34\\1105.3} & \shortstack[r]{34\\1358.6} & \shortstack[r]{34\\1610.2} \\[2pt]
\midrule
Direct & \shortstack[r]{2\\9.7} & \shortstack[r]{3\\28.7} & \shortstack[r]{3\\48.5} & \shortstack[r]{3\\67.3} & \shortstack[r]{3\\86.7} & \shortstack[r]{3\\106.4} & \shortstack[r]{3\\125.8} \\[2pt]
Schema & \shortstack[r]{17\\11.6} & \shortstack[r]{18\\34.4} & \shortstack[r]{18\\57.3} & \shortstack[r]{18\\80.6} & \shortstack[r]{18\\103.4} & \shortstack[r]{18\\126.9} & \shortstack[r]{18\\149.2} \\[2pt]
Reflective & \shortstack[r]{20\\11.1} & \shortstack[r]{20\\32.6} & \shortstack[r]{20\\54.2} & \shortstack[r]{20\\75.8} & \shortstack[r]{20\\97.1} & \shortstack[r]{20\\119.0} & \shortstack[r]{20\\140.3} \\[2pt]
Feedback-Schema & \shortstack[r]{20\\33.4} & \shortstack[r]{21\\100.5} & \shortstack[r]{21\\168.0} & \shortstack[r]{21\\235.4} & \shortstack[r]{21\\300.8} & \shortstack[r]{21\\369.0} & \shortstack[r]{21\\435.2} \\[2pt]
Feedback-Reflective & \shortstack[r]{21\\32.4} & \shortstack[r]{23\\98.3} & \shortstack[r]{23\\161.7} & \shortstack[r]{23\\225.4} & \shortstack[r]{23\\290.6} & \shortstack[r]{23\\355.3} & \shortstack[r]{23\\419.4} \\[2pt]
Plan-and-Execute & \shortstack[r]{10\\13.5} & \shortstack[r]{13\\41.7} & \shortstack[r]{13\\69.3} & \shortstack[r]{13\\96.9} & \shortstack[r]{13\\124.2} & \shortstack[r]{13\\152.4} & \shortstack[r]{13\\180.2} \\[2pt]
ReAct-style & \shortstack[r]{11\\31.9} & \shortstack[r]{13\\95.8} & \shortstack[r]{13\\158.7} & \shortstack[r]{13\\221.0} & \shortstack[r]{14\\284.6} & \shortstack[r]{14\\350.8} & \shortstack[r]{14\\414.6} \\[2pt]
AgentDebug-style & \shortstack[r]{17\\30.1} & \shortstack[r]{21\\67.4} & \shortstack[r]{21\\104.1} & \shortstack[r]{22\\140.5} & \shortstack[r]{22\\176.9} & \shortstack[r]{22\\214.3} & \shortstack[r]{22\\251.8} \\[2pt]
\midrule
\rowcolor{recommitTableHighlight}
ReCommit (LLaDA) & \shortstack[r]{26\\12.6} & \shortstack[r]{32\\38.7} & \shortstack[r]{33\\64.4} & \shortstack[r]{33\\90.3} & \shortstack[r]{33\\115.7} & \shortstack[r]{33\\141.6} & \shortstack[r]{34\\166.6} \\[2pt]
\rowcolor{recommitTableHighlight}
ReCommit (Dream) & \shortstack[r]{25\\12.0} & \shortstack[r]{34\\34.8} & \shortstack[r]{34\\64.4} & \shortstack[r]{36\\93.2} & \shortstack[r]{36\\121.9} & \shortstack[r]{36\\150.5} & \shortstack[r]{36\\179.5} \\[2pt]
\bottomrule
\end{tabular}
\end{table}

\begin{table}[!htbp]
\centering\small
\setlength{\tabcolsep}{4pt}
\renewcommand{\arraystretch}{1.18}
\caption{Recovery--cost results across budgets on Slack (166 episodes). Cells give recovered episodes (top) and mean full-budget repair time in seconds (bottom). Comparison methods use Qwen3-8B unless marked 32B; both shaded ReCommit variants use Qwen3-8B for realization.}
\label{tab:q2-slack}
\begin{tabular}{@{}>{\raggedright\arraybackslash}m{\dimexpr0.30\linewidth-1.75\tabcolsep\relax}>{\raggedleft\arraybackslash}m{\dimexpr0.10\linewidth-1.75\tabcolsep\relax}>{\raggedleft\arraybackslash}m{\dimexpr0.10\linewidth-1.75\tabcolsep\relax}>{\raggedleft\arraybackslash}m{\dimexpr0.10\linewidth-1.75\tabcolsep\relax}>{\raggedleft\arraybackslash}m{\dimexpr0.10\linewidth-1.75\tabcolsep\relax}>{\raggedleft\arraybackslash}m{\dimexpr0.10\linewidth-1.75\tabcolsep\relax}>{\raggedleft\arraybackslash}m{\dimexpr0.10\linewidth-1.75\tabcolsep\relax}>{\raggedleft\arraybackslash}m{\dimexpr0.10\linewidth-1.75\tabcolsep\relax}@{}}
\toprule
\rowcolor{recommitTableHeader}
\textbf{Method} & $B=1$ & $B=3$ & $B=5$ & $B=7$ & $B=9$ & $B=11$ & $B=13$ \\
\midrule
Direct (32B) & \shortstack[r]{46\\37.4} & \shortstack[r]{57\\102.7} & \shortstack[r]{61\\170.4} & \shortstack[r]{62\\245.4} & \shortstack[r]{64\\313.2} & \shortstack[r]{64\\385.8} & \shortstack[r]{65\\454.3} \\[2pt]
Schema (32B) & \shortstack[r]{56\\34.1} & \shortstack[r]{65\\92.7} & \shortstack[r]{66\\153.0} & \shortstack[r]{66\\218.2} & \shortstack[r]{66\\278.4} & \shortstack[r]{66\\338.5} & \shortstack[r]{67\\399.3} \\[2pt]
Feedback-Reflective (32B) & \shortstack[r]{55\\67.6} & \shortstack[r]{70\\190.9} & \shortstack[r]{76\\318.1} & \shortstack[r]{77\\456.6} & \shortstack[r]{81\\588.6} & \shortstack[r]{84\\723.5} & \shortstack[r]{85\\852.7} \\[2pt]
\midrule
Direct & \shortstack[r]{24\\10.5} & \shortstack[r]{26\\31.5} & \shortstack[r]{30\\53.1} & \shortstack[r]{32\\74.0} & \shortstack[r]{33\\94.7} & \shortstack[r]{33\\115.7} & \shortstack[r]{34\\136.7} \\[2pt]
Schema & \shortstack[r]{27\\11.7} & \shortstack[r]{31\\31.8} & \shortstack[r]{33\\51.9} & \shortstack[r]{34\\72.8} & \shortstack[r]{34\\93.8} & \shortstack[r]{35\\114.5} & \shortstack[r]{36\\135.1} \\[2pt]
Reflective & \shortstack[r]{21\\8.8} & \shortstack[r]{27\\27.1} & \shortstack[r]{28\\44.9} & \shortstack[r]{28\\62.3} & \shortstack[r]{28\\78.6} & \shortstack[r]{29\\95.6} & \shortstack[r]{29\\112.8} \\[2pt]
Feedback-Schema & \shortstack[r]{27\\31.3} & \shortstack[r]{32\\90.4} & \shortstack[r]{34\\149.5} & \shortstack[r]{35\\210.3} & \shortstack[r]{35\\269.6} & \shortstack[r]{36\\329.2} & \shortstack[r]{37\\388.2} \\[2pt]
Feedback-Reflective & \shortstack[r]{22\\23.7} & \shortstack[r]{30\\73.2} & \shortstack[r]{31\\122.0} & \shortstack[r]{31\\170.0} & \shortstack[r]{32\\215.6} & \shortstack[r]{32\\261.9} & \shortstack[r]{33\\308.5} \\[2pt]
Plan-and-Execute & \shortstack[r]{27\\12.6} & \shortstack[r]{31\\37.4} & \shortstack[r]{32\\63.5} & \shortstack[r]{35\\88.8} & \shortstack[r]{35\\114.2} & \shortstack[r]{35\\140.0} & \shortstack[r]{35\\163.9} \\[2pt]
ReAct-style & \shortstack[r]{12\\18.8} & \shortstack[r]{19\\56.3} & \shortstack[r]{21\\94.2} & \shortstack[r]{22\\131.4} & \shortstack[r]{26\\169.4} & \shortstack[r]{26\\207.9} & \shortstack[r]{26\\245.2} \\[2pt]
AgentDebug-style & \shortstack[r]{25\\29.2} & \shortstack[r]{34\\61.4} & \shortstack[r]{35\\95.1} & \shortstack[r]{36\\127.7} & \shortstack[r]{36\\160.8} & \shortstack[r]{37\\194.4} & \shortstack[r]{37\\226.8} \\[2pt]
\midrule
\rowcolor{recommitTableHighlight}
ReCommit (LLaDA) & \shortstack[r]{59\\8.7} & \shortstack[r]{75\\30.4} & \shortstack[r]{75\\51.7} & \shortstack[r]{77\\72.8} & \shortstack[r]{79\\94.4} & \shortstack[r]{80\\115.8} & \shortstack[r]{80\\137.7} \\[2pt]
\rowcolor{recommitTableHighlight}
ReCommit (Dream) & \shortstack[r]{60\\8.7} & \shortstack[r]{72\\25.4} & \shortstack[r]{76\\49.1} & \shortstack[r]{76\\73.6} & \shortstack[r]{78\\96.9} & \shortstack[r]{78\\121.2} & \shortstack[r]{79\\145.5} \\[2pt]
\bottomrule
\end{tabular}
\end{table}

\section{Support Search Analysis}
\label{app:arcontrol}

\noindent\textbf{Autoregressive support-proposer control.}
The autoregressive (AR) support-proposer control uses Qwen3-8B to score each operation with a public-context Yes/No probe. It normalizes the largest Yes- and No-token logits into an inclusion score. These scores define binary distributions for support enumeration. The AR proposer uses its own support-ranking rule. Q3 compares recovery between the two complete proposer configurations, with model family and support ordering varying jointly. Both supply supports to the same canonical/expressive schedule and pruning procedure.

Table~\ref{tab:q3} provides the complete per-service recovery counts and paired outcomes for the support-proposer comparison in Figure~\ref{fig:q3paired}(a). ReCommit and the AR-proposer control jointly recover 102 episodes, with 37 recovered only by ReCommit and five only by the control. The aggregate recovery difference is 5.07 percentage points, with a paired task-level bootstrap 95\% confidence interval of $[2.54,7.87]$ points over the AR-proposer control. The shared set accounts for 102 of the AR control's 107 recovered episodes, so the configurations largely overlap on the failures that the control repairs. ReCommit's exclusive recoveries outnumber the control's on each service. Slack contributes the largest net increase, with 23 ReCommit-only recoveries versus one AR-only recovery. These paired outcomes locate the gain in additional coverage beyond the shared recovered set.

\begin{table}[!htb]
\centering
\renewcommand{\arraystretch}{1.13}
\caption{ReCommit and the AR-proposer control at $B=3$. The realizer, public-error pruning, and canonical/expressive schedule are fixed. Each proposer retains its own support-ordering rule. The last two columns count episodes recovered by one configuration but missed by the other. Higher recovery counts within each row are \textbf{bolded}.}
\label{tab:q3}
\small
\setlength{\tabcolsep}{6pt}
\begin{tabular}{@{}>{\raggedright\arraybackslash}p{\dimexpr0.170000\linewidth-1.666667\tabcolsep\relax}>{\raggedleft\arraybackslash}p{\dimexpr0.130000\linewidth-1.666667\tabcolsep\relax}>{\raggedleft\arraybackslash}p{\dimexpr0.180000\linewidth-1.666667\tabcolsep\relax}>{\raggedleft\arraybackslash}p{\dimexpr0.150000\linewidth-1.666667\tabcolsep\relax}>{\raggedleft\arraybackslash}p{\dimexpr0.180000\linewidth-1.666667\tabcolsep\relax}>{\raggedleft\arraybackslash}p{\dimexpr0.190000\linewidth-1.666667\tabcolsep\relax}@{}}
\toprule
\multirow[c]{2}{*}{\textbf{Service}} & \multirow[c]{2}{*}{\textbf{Episodes}} & \multicolumn{2}{c}{\textbf{Recovered episodes} $\uparrow$} & \multicolumn{2}{c}{\textbf{Exclusive recoveries}} \\
\cmidrule(lr){3-4}\cmidrule(lr){5-6}
& & \shortstack{\textbf{AR-proposer}\\\textbf{control}} & \textbf{ReCommit} & \shortstack{\textbf{ReCommit}\\\textbf{only}} & \shortstack{\textbf{AR-proposer}\\\textbf{control only}} \\
\midrule
Box & 129 & 16 & \textbf{23} & 9 & 2 \\
Calendar & 180 & 7 & \textbf{9} & 2 & 0 \\
Linear & 156 & 31 & \textbf{32} & 3 & 2 \\
Slack & 166 & 53 & \textbf{75} & 23 & 1 \\
\midrule
\rowcolor{recommitTableHighlight}
\textcolor{recommitTableAccent}{\textbf{Total}} & 631 & 107 & \textbf{139} & 37 & 5 \\
\bottomrule
\end{tabular}
\end{table}

\subsection{Masked-Slot Sensitivity}
\label{app:slots}

Table~\ref{tab:slots} supplements Figure~\ref{fig:q3paired}(b) with service-level recovery, repair time, and results without pruning. The aggregate optimum does not coincide with every service's best setting: Calendar recovers 17 episodes at $M=20$, compared with nine at the shared default $M=8$, while recovery decreases on the other three services. Increasing the slot count therefore changes service-level coverage without consistently expanding it. Larger slot counts also increase repair time, and the widening gap between recovery with and without pruning indicates a greater contribution from public-error pruning. The shared default balances aggregate recovery and cost across the evaluated services.

\begin{table}[!htb]
\centering
\renewcommand{\arraystretch}{1.13}
\small
\caption{Effect of the masked-slot count $M$ on recovery and cost at $B=3$. Service columns give recovered episodes with public-error pruning; With-P and Without-P give totals with and without pruning. Time is mean full-budget repair time with pruning (s). The shared default $M=8$ is shaded; the best aggregate recoveries are bolded.}
\label{tab:slots}
\begin{tabular}{@{}>{\raggedleft\arraybackslash}p{\dimexpr0.055000\linewidth-1.750000\tabcolsep\relax}>{\raggedleft\arraybackslash}p{\dimexpr0.100000\linewidth-1.750000\tabcolsep\relax}>{\raggedleft\arraybackslash}p{\dimexpr0.140000\linewidth-1.750000\tabcolsep\relax}>{\raggedleft\arraybackslash}p{\dimexpr0.100000\linewidth-1.750000\tabcolsep\relax}>{\raggedleft\arraybackslash}p{\dimexpr0.100000\linewidth-1.750000\tabcolsep\relax}>{\raggedleft\arraybackslash}p{\dimexpr0.170000\linewidth-1.750000\tabcolsep\relax}>{\raggedleft\arraybackslash}p{\dimexpr0.180000\linewidth-1.750000\tabcolsep\relax}>{\raggedleft\arraybackslash}p{\dimexpr0.155000\linewidth-1.750000\tabcolsep\relax}@{}}
\toprule
\rowcolor{recommitTableHeader}
$M$ & \multicolumn{4}{c}{\textbf{Recovered episodes by service}} & \multicolumn{2}{c}{\textbf{Total recovered}} & \textbf{Time} \\
\cmidrule(lr){2-5}\cmidrule(lr){6-7}
\rowcolor{recommitTableHeader}
& Box & Calendar & Linear & Slack & With-P & Without-P & (s) \\
\midrule
4 & 11 & 5 & 29 & 67 & 112 & 110 & 25.99 \\
\rowcolor{recommitTableHighlight}
8 & 23 & 9 & 32 & 75 & \textbf{139} & \textbf{128} & 33.60 \\
12 & 22 & 9 & 29 & 71 & 131 & 92 & 55.20 \\
16 & 18 & 12 & 26 & 65 & 121 & 71 & 58.37 \\
20 & 16 & 17 & 27 & 65 & 125 & 62 & 63.44 \\
\bottomrule
\end{tabular}
\end{table}